\documentclass[review]{elsarticle}

\usepackage{geometry}
\usepackage{pdflscape}
\usepackage{lineno,hyperref}
\usepackage{xcolor}
\usepackage{scalerel,verbatimbox}
\modulolinenumbers[5]

\def\mycircle{\scalerel*{\addvbuffer[-.15pt -.6pt]{$\circ$}}{\blacksquare}}

\newcommand{\bluecircle}{\textcolor{blue}{\mycircle}}
\newcommand{\redsquare}{\textcolor{red}{\blacksquare}}

\journal{Pattern Recognition Journal}

\usepackage[utf8]{inputenc}
\usepackage[T1]{fontenc}
\usepackage{amssymb}
\usepackage[cmex10]{amsmath}
\usepackage{algorithm,algorithmicx,algpseudocode}

\usepackage{mdwmath}
\usepackage{mdwtab}
\usepackage[tight,footnotesize]{subfigure}
\usepackage{url}
\usepackage{natbib}
\usepackage{tabularx}
\usepackage{epsfig}
\usepackage{epstopdf}
\usepackage{tikz}
\usetikzlibrary{arrows,calc,intersections}

\makeatletter
\newcounter{phase}[algorithm]
\newlength{\phaserulewidth}
\newcommand{\setphaserulewidth}{\setlength{\phaserulewidth}}

\makeatother

\setphaserulewidth{.7pt}
	  
\begin{document}

\begin{frontmatter}

\title{
    FIRE-DES++: Enhanced Online Pruning of Base Classifiers\\
    for Dynamic Ensemble Selection
}

\author[ets]{Rafael M. O. Cruz}
\ead{rafaelmenelau@gmail.com}

\author[cin]{Dayvid V. R. Oliveira\corref{mycorrespondingauthor}}
\cortext[mycorrespondingauthor]{Corresponding author}
\ead{dvro@cin.ufpe.br}

\author[cin]{George D. C. Cavalcanti}
\ead{gdcc@cin.ufpe.br}

\author[ets]{Robert Sabourin}
\ead{robert.sabourin@etsmtl.ca}

\address[cin]{Centro de Inform\'atica - Universidade Federal de Pernambuco, Brazil}
\address[ets]{\'Ecole de Technologie Sup\'erieure - Universit\'e du Qu\'ebec}

\begin{abstract}
Dynamic Ensemble Selection (DES) techniques aim to select
one or more competent classifiers for the classification
of each new test sample. Most DES techniques estimate the competence of classifiers
using a given criterion over the region of competence of
the test sample, usually defined as the set of nearest
neighbors of the test sample in the validation set.
Despite being very effective in several classification tasks,
DES techniques can select classifiers that classify all samples
in the region of competence as being from the same class.
The Frienemy Indecision REgion DES (FIRE-DES) tackles
this problem by pre-selecting classifiers that correctly
classify at least one pair of samples from different classes
in the region of competence of the test sample.
However, FIRE-DES applies the pre-selection for the classification of a test sample
if and only if its region of competence is composed of samples
from different classes (indecision region),
even though this criterion is not reliable for determining if a test sample
is located close to the borders of classes (true indecision region)
when the region of competence is obtained using classical nearest neighbors approach.
Because of that, FIRE-DES mistakes noisy regions for true indecision regions,
leading to the pre-selection of incompetent classifiers,
and mistakes true indecision regions for safe regions,
leaving samples in such regions without any pre-selection.
To tackle these issues, we propose the FIRE-DES++,
an enhanced FIRE-DES that removes noise and reduces the overlap of classes in the validation set;
and defines the region of competence using an equal number of samples
of each class, avoiding selecting a region of competence with samples
of a single class.
Experiments are conducted using FIRE-DES++ with 8 different dynamic
selection techniques on 64 classification datasets.
Experimental results show that FIRE-DES++ increases the classification
performance of all DES techniques considered in this work,
outperforming FIRE-DES with 7 out of the 8 DES techniques,
and outperforming state-of-the-art DES frameworks.

\end{abstract}

\end{frontmatter}

\section{Introduction}

Dynamic Ensemble Selection (DES) has become an important research
topic in the last few years \cite{cruz2018dynamic}.
Given a test sample and a pool of classifiers, DES techniques
select one or more competent classifiers for the classification
of that test sample.
The most important part in DES techniques is how to evaluate the
competence level of each base classifier for the classification
of a given test sample \cite{cruz2016prototype}.
In general, DES techniques evaluate the competence level of base
classifiers for the classification of a test sample, $x_{query}$,
based on the performance of the base classifier in a local region surrounding the test sample, named region of competence.
Most DES techniques define the region of competence of test samples
using the K-Nearest Neighbors of the test sample in the validation
set, we refer to this validation set as the dynamic selection dataset ($D_{SEL}$) \cite{dcs:2014}.

Despite being very effective in several classification tasks,
DES techniques can select classifiers that classify all samples
in the region of competence of a test sample to the same class,
even when the test sample is located close to a decision border, having neighbors belonging to different classes
(indecision region) \cite{dfp:2017}.

\begin{figure}[ht]
	\centering
	\includegraphics[scale=0.3]{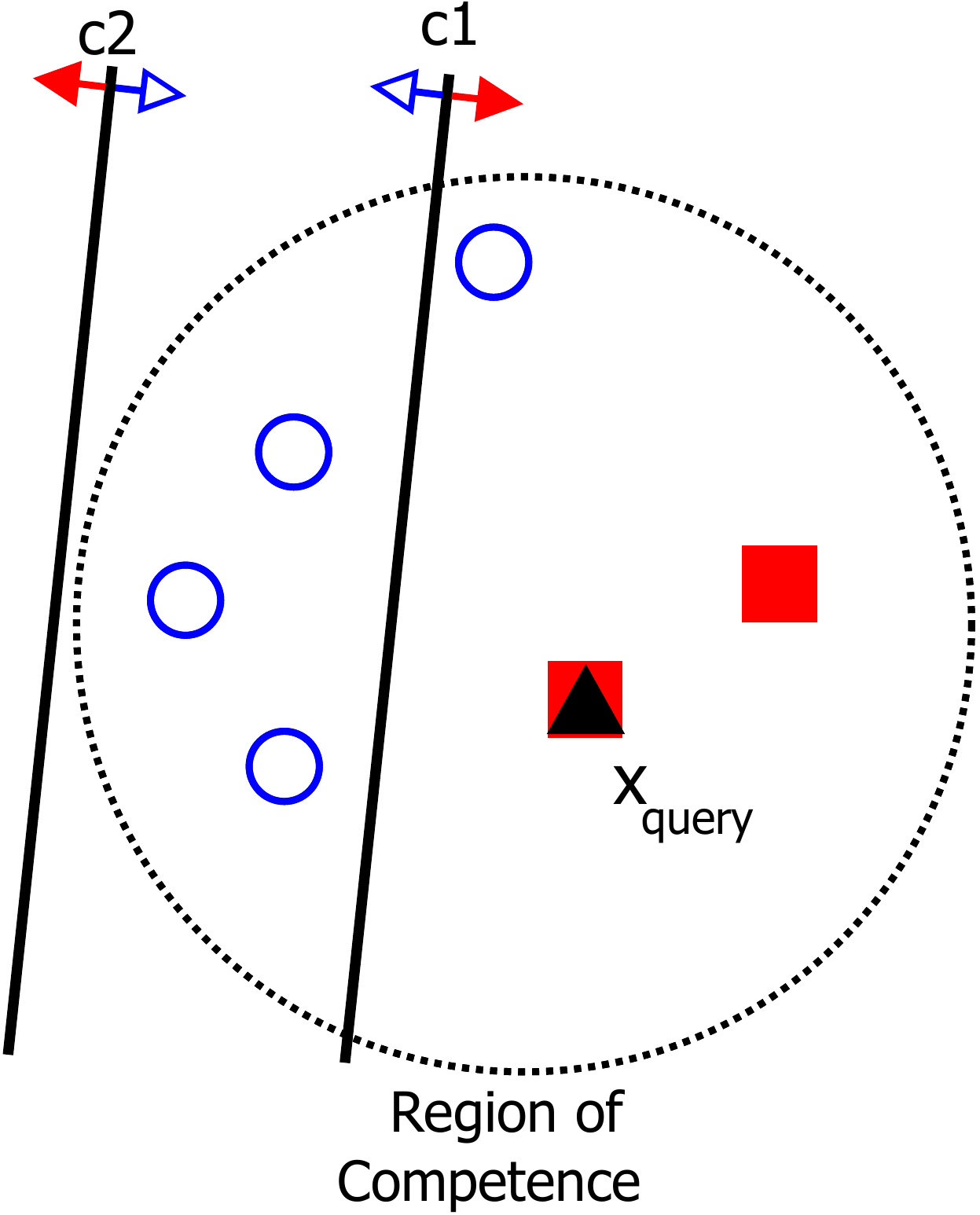}
	\caption{$c1$ crosses the region of competence and predict the correct label for samples from different classes, while $c2$ can only correctly classify the samples belonging to the blue class.}
	\label{fig:decision_boundaries}
\end{figure}

Figure~\ref{fig:decision_boundaries} represents a query sample, $x_{query}$, located in a indecision region.
In this example, the decision boundary of classifier $c1$ crosses the region of competence of $x_{query}$, and it predicts different class labels for the samples belonging to this region. It also correctly classifies at least one sample from each class. On the other hand, $c2$ does not cross the region
of competence of $x_{query}$. However, since it correctly classifies the same number of samples as $c1$, a DES algorithm could select $c2$ as a local competent classifier, instead of $c1$, misclassifying the query.

To deal with this issue, Oliveira et al. \cite{dfp:2017} proposed the
Frienemy Indecision Region Dynamic Ensemble Selection (FIRE-DES),
a DES framework that pre-selects classifiers with decision boundaries
crossing the region of competence when the test sample is located in an indecision region.
Given a test sample $x_{query}$, FIRE-DES decides if it
is located in an indecision region. If so, it uses the
Dynamic Frienemy Pruning (DFP) to pre-select classifiers with decision boundaries crossing the region
of competence of $x_{query}$. Then, only the pre-selected pool is passed down to a DES technique to select
the final ensemble of classifiers.

However, the FIRE-DES does not consider whether or not the region of competence is a good representation of the type of region
in which the test sample is located. For instance, the FIRE-DES can mistake a safe region as being an indecision region due to the presence of noise in $D_{SEL}$. In this case, the DFP can remove local competent classifiers from the pool as they do not correctly classify the noise instance, leaving only the base classifiers that modeled the noise in the local region for the DES step. 

In addition, when dealing with small sized datasets, some regions of the feature space may not be well populated. In such cases, the region of competence of $x_{query}$ can contain samples belonging to a single class (safe region) even though $x_{query}$ may be located close to the class borders (true indecision region). In such cases, the FIRE-DES algorithm will mistake that $x_{query}$ is located in a safe region.  Hence, the DFP algorithm will not be employed to remove incompetent classifiers. However, the query is located in a true indecision region since it is close to the
decision border of classes, regardless of the classes represented in its region of competence.

In this paper, we propose the FIRE-DES++, an enhanced FIRE-DES framework that tackles the 
noise sensitivity and indecision region restriction drawbacks of the previous framework. 
The main differences between the FIRE-DES++ to the original version are:
(1) The FIRE-DES++ applies a prototype selection (PS) technique in order to remove noise from the validation set ($D_{SEL}$). 
Hence, the FIRE framework will not mistake a noisy region for an indecision region when estimating the regions of competence. 
(2) During the test phase, the FIRE-DES++ employs a K-Nearest Neighbors Equality (KNNE)~\cite{knne:2011} to define the region
of competence. The KNNE is a variation of the KNN technique which selects the same amount of samples from each class. 
By using the KNNE, test instances that are located close to the decision borders
(in a true indecision region) will never be mistaken as belonging to a safe region since its region
of competence will always be composed of samples from different classes. Thus, solving the indecision region 
restriction drawback of the FIRE-DES framework. Like FIRE-DES, FIRE-DES++ can be used with any dynamic selection technique based on the nearest neighbors to estimate the competence level of base classifiers. 

The experiments were conducted over 64 datasets from the Knowledge Extraction based on Evolutionary Learning (KEEL) repository \cite{keel}. We evaluated FIRE-DES++ on 8 dynamic selection techniques: Overall Local Accuracy (OLA)~\cite{dcs_la:1996},
Local Class Accuracy(LCA)~\cite{dcs_la:1996}, A Priori selection~\cite{dcs_la:1999}, A Posteriori selection~\cite{dcs_la:1999}, 
Multiple Classifier Behavior (MCB)~\cite{mcb:2001}, Dynamic Selection KNN~\cite{dsknn:2006} and the K-Nearest
Oracles Union (KNU) and Eliminate (KNE)~\cite{des:2008}. We also compared FIRE-DES++ with the better performing dynamic selection technique according to a recent survey~\cite{cruz2018dynamic}: Randomized Reference Classifier (RRC) \cite{rrc:2011}, META-DES \cite{metades:2015}, and META-DES.Oracle \cite{metadesoracle:2017} as well as several static ensemble approaches. 

This paper is organized as follows: Section 2 presents the problem statement,
Section 3 presents the proposed framework, Section 4 presents the experimental study,
and Section 5 concludes the paper.

\section{Problem Statement}
\label{sec:ps}

\subsection{FIRE-DES}

The Frienemy Indecision Region Dynamic Ensemble Selection (FIRE-DES) framework works as an online pruning mechanism to pre-select base classifiers before applying the dynamic ensemble selection techniques. Given a new input query to the system, $x_{query}$, the FIRE-DES framework analyze its region of competence to decide whether or not it is located in an indecision region (region of competence with samples from different classes). If the sample is located in a safe region, i.e., the whole region of competence is composed of samples belonging to the same class, all base classifiers are passed down to the dynamic selection technique. However, when the query is located on an indecision region, the framework applies the Dynamic Frienemy Pruning (DFP) technique to pre-select base classifiers that are able to correctly classify at least a pair of samples belonging to different classes in the region of competence. This pair of samples is called frienemy. Two instances $x_a$ and $x_b$ are considered frienemies if they are located in the region of competence of $x_{query}$, and have different class labels. 

Ideally, a local competent classifier would be able to distinguish all frienemies pair in the region of competence, thus being able to separate between the two classes locally. The DFP is applied to pre-select only the base classifiers that correctly classify at least one pair of frienemies. Then, only the pre-selected base classifiers are passed down to the DES algorithm for the competence estimation and classification. In the example presented in Fig~\ref{fig:decision_boundaries}, the DFP would remove $c2$ since it does not correctly classify a single pair of frienemies. That way, although $c1$ and $c2$ may have the same local competence level, $c2$ would not be taken into consideration by the DS algorithm. Hence, the $c1$ would be selected predicting the correct label of the query. In a case where no base classifier correctly classifies a single pair of frienemies, all base classifiers are considered for competence estimation. 

Although the FIRE-DES framework can be used to significantly improve the performance of several DES techniques \cite{dfp:2017}, it suffers from two main drawbacks: the noise sensitivity, and indecision region restriction.

\subsection{Drawback 1: Noise Sensitivity}
\label{subsec:p1}

\begin{figure*}[t]
	\centering
	\subfigure[
	Toy problem of noisy region of competence (A, B, C, and N),
	the markers {\Large$\textcolor{blue}{\circ}$} (A, B, C, and D) and
	$\textcolor{red}{\blacksquare}$ (N, E, and F)
	are samples of different classes,
	the sample labeled $N$ is a noisy sample.
	]{
		\includegraphics[scale=0.3]{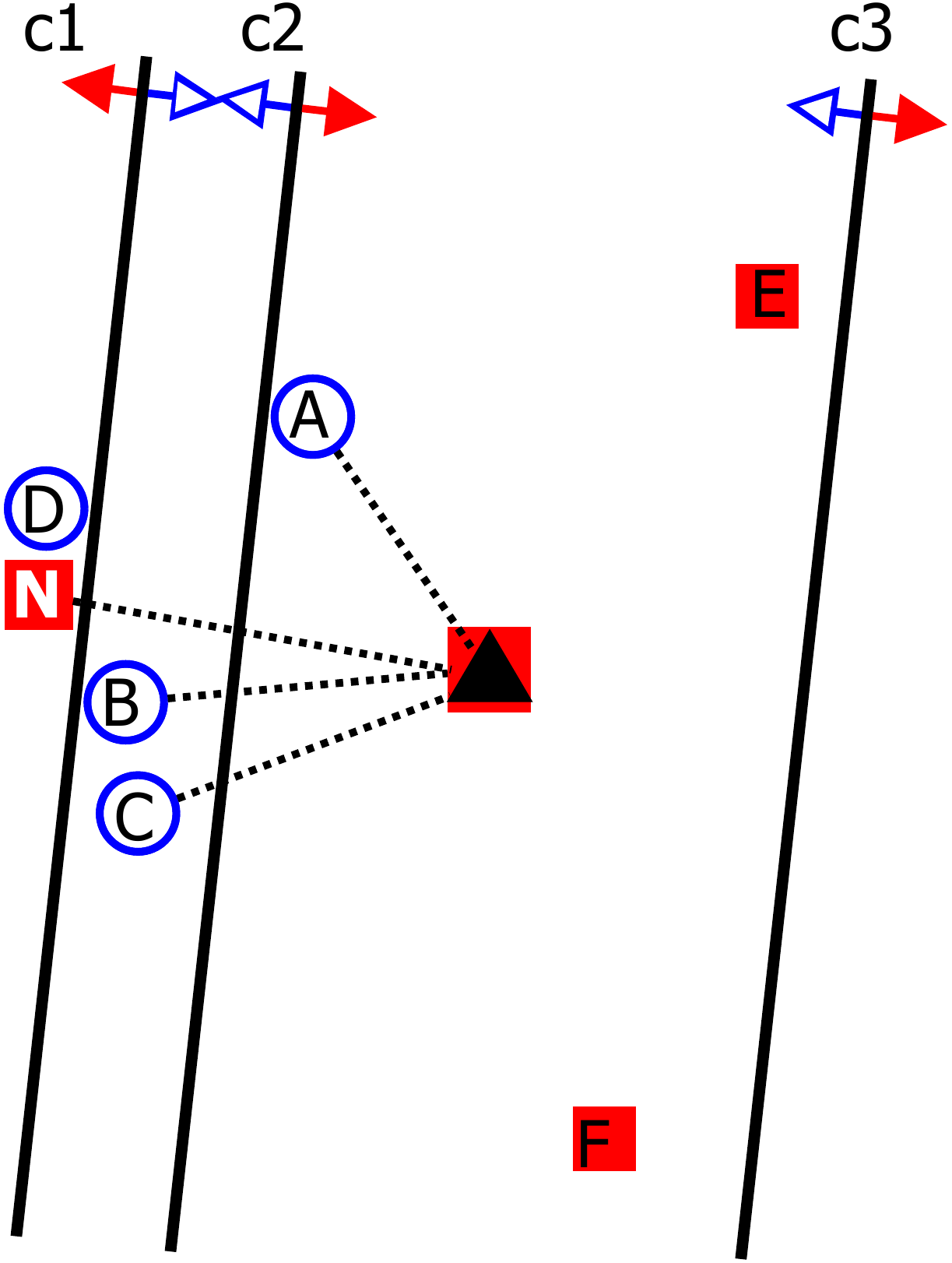}
		\label{fig:ps1}
	}
	\hspace*{2cm}
	\subfigure[
	Toy problem of a test sample $\blacktriangle$ and  a filtered
	- noisy sample N was removed - region of competence
	(A, B, C, and D),
	the markers {\Large$\textcolor{blue}{\circ}$} (A, B, C, and D) and
	$\textcolor{red}{\blacksquare}$ (E, and F)
	are samples of different classes.
	]{
		\includegraphics[scale=0.3]{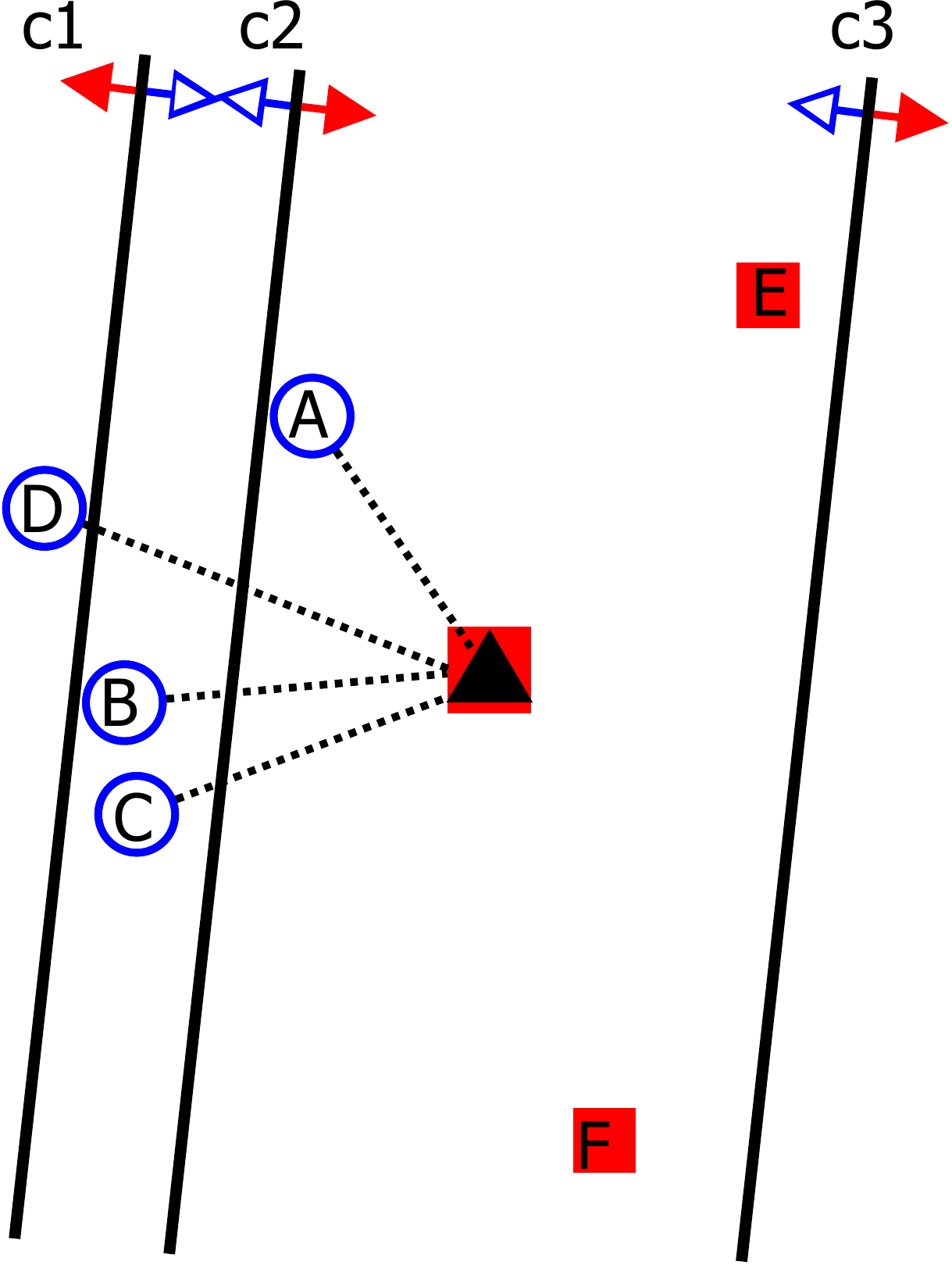}
		\label{fig:ps2}
	}
	\caption{
		DES applied to the classification of a test sample $\blacktriangle$
		of class $\textcolor{red}{\blacksquare}$.
		The continuous straight lines are the decision boundaries of
		classifiers $c1$, $c2$, and $c3$,
		the markers {\Large$\textcolor{blue}{\circ}$} (A, B, C, and D) and
		$\textcolor{red}{\blacksquare}$ (N, E, and F)
		are samples of different classes,
		N is a noisy sample,
		and samples connected to the test sample by a dotted line
		define the region of competence of the test sample.
	}
	\label{fig:ps}
\end{figure*}

The noise sensitivity drawback is important because DES techniques are highly
sensitive to noise, outliers, and high level of overlap between classes in $D_{SEL}$ \cite{cruz2016prototype, dsfa:2011}. Figure \ref{fig:ps1} shows a test sample ($\blacktriangle$)
with true class $\textcolor{red}{\blacksquare}$ located in
a noisy region, and three classifiers
$c1$, $c2$, and $c3$. In this figure, the region of competence ($\Psi$)
of the test sample is composed of the samples A, B, C, and N
(sample $N$ is noise).
In the example from Figure \ref{fig:ps1},
the classifier $c1$ correctly classifies 4 samples in $\Psi$
(A, B, C, and the noise instance N),
the classifier $c2$ correctly classifies 2 samples in $\Psi$
(B, and C),
and the classifier $c3$ correctly classifies 3 samples in $\Psi$
(A, B, and C). 



The Overall Local Accuracy (OLA) \cite{dcs_la:1996} DES technique
estimates the competence of classifiers using their accuracy in
the region of competence, that is, the more samples a classifier
correctly classifies, the more competent it is. OLA selects only
the most competent classifier for the classification of the
test sample.

In Figure \ref{fig:ps1}, OLA selects $c1$, the classifier that
correctly classify most samples in $\Psi$,
even though $c1$ was only considered the best because of a noisy
sample (N). This selection leads to the misclassification of the
test sample as $\bluecircle$.
Also in this example, the FIRE-DES will mistake the noisy region
(region with noisy samples) for an indecision region (region
composed of samples from different classes), and pre-select
classifiers that correctly classify at least one pair of samples
from different classes (frienemies), in this case $c1$,
also misclassifying the test sample as $\bluecircle$.

\subsection{Drawback 2: Indecision Region Restriction}
\label{subsec:p2}

Figure \ref{fig:ps2} shows the scenario from Figure \ref{fig:ps1}
without the noisy sample $N$.
Figure \ref{fig:ps2} shows a test sample ($\blacktriangle$)
with true class $\textcolor{red}{\blacksquare}$ located in a
true indecision region (close to the borders),
and three classifiers $c1$, $c2$, and $c3$.
In this figure, the region of competence ($\Psi$)
of the test sample is composed of the samples A, B, C, and D all from
class $\bluecircle$.
In the example from Figure \ref{fig:ps2},
the classifier $c1$ correctly classify 3 samples in $\Psi$
(A, B, and C),
the classifier $c2$ correctly classify 2 samples in $\Psi$
(B, and C),
and the classifier $c3$ correctly classify 4 samples in $\Psi$
(A, B, C, and D). 

In Figure \ref{fig:ps2}, OLA selects the classifier that correctly
classify the most samples in $\Psi$, that is, $c3$, even though
$c3$ classify all samples in the region of competence of the
test sample as being from the same class $\bluecircle$,
misclassifying the test sample.

In the example from Figure \ref{fig:ps2}, the FIRE-DES does not
apply the DFP because it considers $x_{query}$ as being located
in a safe region, even though it is located in a true indecision region. 
Therefore, FIRE-DES with OLA also misclassifies the test sample as being from the class $\bluecircle$.
This scenario is very likely to happen when dealing with small sized as well as imbalanced datasets, in which
one of the classes may not contain enough examples in the local region. 

\section{The proposed framework}

\begin{figure*}[htbp]
	\begin{center}
		\resizebox{1.0\textwidth}{!}{%
			\includegraphics[scale=0.5]{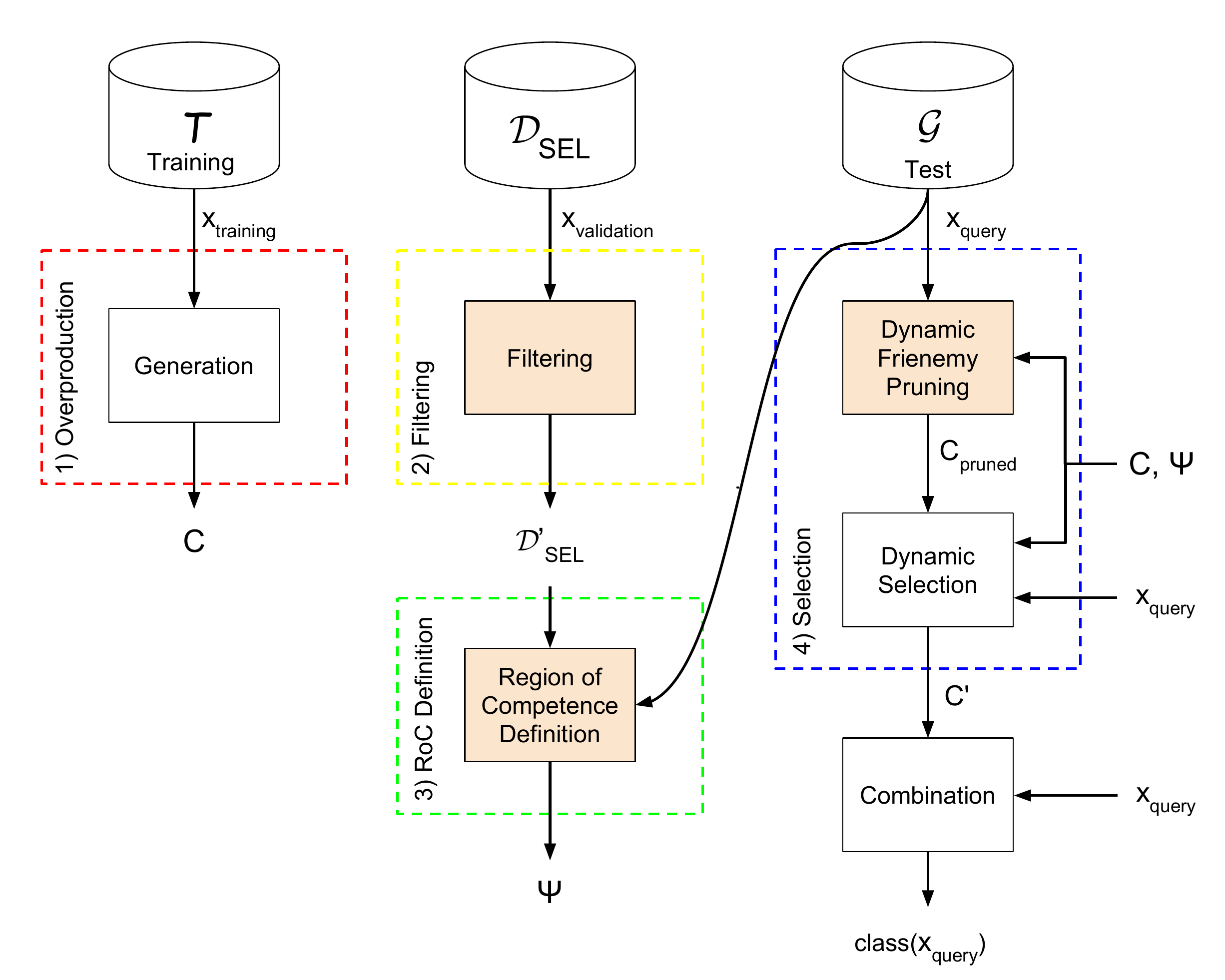}
		}
		\caption{
			Overview of FIRE-DES++,
			where $\mathcal{G}$ is the test set,
			$x_{query}$ is the test sample,
			$\mathcal{T}$ is the training set,
			\textit{Generation} is a ensemble generation process (i.e. Bagging)
			used to generate the pool of classifiers $C$,
			$\mathcal{D}_{SEL}$ is the validation set, 
			\textit{Filtering} is the process of filtering $\mathcal{D}_{SEL}$ using
			a prototype selection algorithm which results in the improved validation
			set $\mathcal{D}_{SEL}'$,
			\textit{Region of competence definition (RoCD)} is the process of selecting the
			region of competence $\Psi$ of $x_{query}$ with size $K$,
			\textit{Dynamic Frienemy Pruning} is the Dynamic Frienemy Pruning (DFP) step,
			\textit{Dynamic Selection} is the Dynamic Selection step,
			$C_{pruned}$ is the set of pre-selected classifiers, 
			$C'$ is the ensemble of selected classifiers for the
			classification of $x_{query}$,
			\textit{Combination} is a combination rule,
			and \textit{class($x_{query}$)} is the final classification of 
			$x_{query}$.
		}
		\label{fig:e2f_overview}
	\end{center}
\end{figure*}

In this section, we propose an enhanced Frienemy Indecision Region Dynamic Ensemble Selection
(FIRE-DES++). FIRE-DES++ is divided into four phases (Figure \ref{fig:e2f_overview}): overproduction, filtering, region of competence definition and selection. The main differences between the original FIRE-DES framework and the proposed FIRE-DES++ are the addition of the filtering phase to deal with the noise sensitivity drawback, and the region of competence definition phase, in which the KNN-Equality is applied to guarantee that all classes are represented in the region of competence. Algorithms~\ref{alg:fire++training} and ~\ref{alg:fire++test} present the training and test stages of the FIRE-DES++ framework, respectively.

\begin{enumerate}
	\item \textbf{Overproduction phase}, where the pool of classifiers
	$C$ is generated using the training set ($\mathcal{T}$).
	The overproduction phase is performed only once in the
	training stage.
	
	\item \textbf{Filtering phase}, where
	a Prototype Selection (PS) \cite{ps-taxonomy:2012} technique is applied
	to the validation set $\mathcal{D}_{SEL}$, removing noise and outliers, and reducing
	the level of overlap between classes in $\mathcal{D}_{SEL}$.
	The improved validation set is named $\mathcal{D}_{SEL}'$.
	The filtering phase is performed only once in the training stage.
	
	\item \textbf{Region of competence definition (RoCD) phase},
	there the framework defines the region of competence ($\Psi$)
	using the K-Nearest Neighbors Equality (KNNE) \cite{knne:2011}
	to select samples from the improved validation set $\mathcal{D}_{SEL}'$.
	The KNNE is a nearest neighbor approach that selects an equal
	number of samples from each class, avoiding the definition of
	a region of competence with samples of a single class.
	The RoCD phase is performed in the testing stage for each new
	test sample.
	
	\item \textbf{Selection phase},
	where the ensemble of classifiers
	for the classification of each new test sample is selected.
	Given a test sample $x_{query}$,
	this phase pre-selects base classifiers with decision
	boundaries crossing the region of competence of $x_{query}$
	($C_{pruned}$), if such classifier exists,
	using the Dynamic Frienemy Pruning (DFP) \cite{dfp:2017}.
	The DFP pre-selects classifiers that correctly classify at least
	one pair of samples from different classes ("frienemies") in the
	region of competence.
	The DFP avoids the selection of classifiers that classify all
	samples in the region of competence as being from the same class.
	After the pre-selection, any DES technique is applied to
	perform to select the final ensemble of classifiers ($C'$).
	Finally, the framework uses a combination rule to combine the
	predictions of the selected classifiers into a single prediction.
	
\end{enumerate}

\begin{algorithm}[htbp]
	\caption{FIRE-DES++ training stage}
	\small
	\begin{algorithmic}[1]
		\Require Training data, $\mathcal{T}$
		\Require Validation data, $D_{SEL}$ 
		
		\State $\mathcal{C} =  PoolGeneration(\mathcal{T})$ \Comment{Generate a pool of classifiers based on the training dataset} 
		
		
		\State $\mathcal{D}_{SEL}' = PrototypeSelection(\mathcal{D}_{SEL})$  \Comment{Apply prototype selection to modify the distribution of $D_{SEL}$} 
		
		\State \Return  $\mathcal{C}$, $\mathcal{D}_{SEL}'$
		
	\end{algorithmic}
	\label{alg:fire++training}
\end{algorithm}

\setlength{\textfloatsep}{1pt}
\begin{algorithm}[htbp]
	\caption{FIRE-DES++ testing stage}
	\small
	\begin{algorithmic}[1]
		
		\Require $x_{query}$: Input sample
		\Require $\mathcal{C}$: pool of classifiers
		\Require $\mathcal{D}_{SEL}'$: Filtered dynamic selection dataset
		
		\State $\Psi$ = KNN-Equality$(\mathcal{D}_{SEL}', x_{query})$  \Comment{Get the region of competence $\Psi$} 
		
		\State $C_{pruned}$ $\gets DFP(\Psi, \mathcal{C})$  \Comment{Apply the DFP pruning} 
		
		\State $C' =  DES(\Psi, C_{pruned})$ \Comment{Perform dynamic ensemble selection over the pruned pool} 
		
		\State $class(x_{query}) = Combination(C', x_{query})$ \Comment{Predicting using the selected ensemble $C'$}
		
		\State \Return $class(x_{query})$
	\end{algorithmic}
	\label{alg:fire++test}
\end{algorithm}

In Figure \ref{fig:e2f_overview},
$\mathcal{T}$ is the training set,
\textit{Generation} is an ensemble generation process (i.e. Bagging \cite{bagging:1996}),
and $C$ is the generated pool of classifiers;
$\mathcal{G}$ is the test set,
$x_{query}$ is the test sample;
$\mathcal{D}_{SEL}$ is the validation set,
\textit{Filtering} is the process of filtering $\mathcal{D}_{SEL}$ using
a prototype selection algorithm which results in the improved validation
set $\mathcal{D}_{SEL}'$,
\textit{Region of Competence Definition} is the process of selecting
the region of competence of $x_{query}$ using the filtered validation set $\mathcal{D}_{SEL}'$,
$\Psi$ is the region of competence of $x_{query}$;
\textit{Dynamic Frienemy Pruning} is the Dynamic Pruning step,
$C_{pruned}$ is the pre-selected ensemble of classifiers,
\textit{Dynamic Selection} is the Dynamic Selection step;
$C'$ is the ensemble of selected classifiers,
\textit{Combination} is the process of combining the prediction
of the classifiers in $C'$,
and \textit{class($x_{query}$)} is the final prediction of $x_{query}$.

The phases of FIRE-DES++ complement each other as
the filtering phase tackles the noise sensitivity drawback,
removing noise and reducing the level of overlap between classes;
the region of competence definition phase tackles the indecision region restriction
drawback, as it ensures that all classes are represented in
the region of competence of the test sample;
and, finally, the selection phase pre-selects classifiers with
decision boundaries crossing the region of competence, without
having to consider the effect of noise
(since noise is removed in the filtering phase),
or deciding if a test sample is located in an indecision
region or not (as the region of competence definition phase always
selects regions of competence composed of samples of different classes). The phases of FIRE-DES++ are detailed in the following subsections.

\subsection{Overproduction}
\label{sec:overproduction}

The overproduction phase uses any ensemble generation technique
to generate the pool of classifiers $C$ trained with the training
set $\mathcal{T}$.
Since the focus of this work is on dynamic selection, the
Bagging technique \cite{bagging:1996} \cite{bagging:1998} is used to generate
the pool of classifiers, following the approach used in \cite{dfp:2017}. 

\subsection{Filtering phase}
\label{sec:filtering}

The filtering phase tackles the noise sensitivity drawback
(Section \ref{subsec:p1}), as removing noise from $\mathcal{D}_{SEL}$,
preventing FIRE-DES from estimating the competence level of base 
classifiers using noisy data. This step is conducted by applying a PS technique
to the validation set ($\mathcal{D}_{SEL}$), resulting in an
improved validation set ($\mathcal{D}_{SEL}'$) with less noise, and
less overlap between classes.

In \cite{ps-taxonomy:2012}, the authors presented a taxonomy
of prototype selection, classifying prototype selection techniques
into three categories:
(1) Condensation techniques, that remove samples in the
center of classes, maintaining the borderline samples.
(2) Edition techniques, that remove sample in the borders
of classes,
maintaining safe samples (located in the center of classes).
(3) Hybrid techniques, that combine condensation and edition
approaches.

We expect the filtering phase to cause a high performance gain to
the FIRE-DES++ framework, as in \cite{cruz2016prototype}, the authors
show that state-of-the-art techniques fail to obtain a good 
approximation of the decision boundaries of classes
when noise is added to $\mathcal{D}_{SEL}$, and also demonstrate that using 
PS increases the classification performance of DES techniques.

Two PS techniques are considered: the Relative Neighborhood Graph (RNG) \cite{rng:1997}
and the Edited Nearest Neighborhood (ENN) \cite{enn:1972}. These two PS techniques were the best approaches
for dynamic selection purposes according to \cite{dsfa2:2017}. Furthermore, since our experimental study is focused on small datasets with different levels of class imbalance, only samples of the majority class are removed
from the validation set. Therefore, they also help to alleviate class imbalance problems when performing dynamic selection~\cite{roy2018study}.

\subsubsection{Relative Neighborhood Graph (RNG)}

The RNG technique uses the concept of Proximity Graph (PG)
to select prototypes. RNG builds a PG, G = (V, E), in which
the vertices are samples (V = $\mathcal{D}_{SEL}$) and the
set of edges E contains an edge connecting two samples $(x_i, x_j)$
if and only if $(x_i,x_j)$ satisfy the neighborhood criterion
in Equation \ref{eq:rng}:

\begin{equation}
\label{eq:rng}
\begin{aligned}
(x_i,x_j) \in E \Leftrightarrow dist(x_i,x_j) \le max(dist(x_i,x_k), dist(x_j, x_k))\\\forall x_k \in X, k = i, j
\end{aligned}
\end{equation}

\noindent where $dist$ is the Euclidean distance between two samples,
and $X$ is the validation set $D_{SEL}$.
The corresponding geometric is defined as the disjoint intersection
between two hyperspheres centered in $x_i$ and $x_j$, and radius
equal to $dist(x_i,x_j)$. Two samples are relative neighbors if
and only if this intersection does not contain any other sample
from $\mathcal{D}_{SEL}$. The relative neighborhood of a sample is
the set of all its relative neighbors.
After building the PG and defining all graph neighbors,
all samples with class label different from the majority of
their respective relative neighbors are removed from $\mathcal{D}_{SEL}$.

Algorithm \ref{alg:rng} presents the pseudo-code of the RNG
technique used in this work.
Given the validation set $\mathcal{D}_{SEL}$,
all samples are added in the filtered validation set
$\mathcal{D}_{SEL}'$ (Line 1),
and the proximity graph of the samples in $\mathcal{D}_{SEL}$
are stored in $PG$ (Line 2).
Now, for each sample $x_i \in \mathcal{D}_{SEL}$,
the relative neighbors ($RN$) of $x_i$ are selected,
and, if the most common class label in $RN$ is different
from the class label of $x_i$, and $x_i$ is not from the
minority class, 
$x_i$ is removed from the filtered validation set
$\mathcal{D}_{SEL}'$ (Line 3 - 10).
Finally, the filtered validation set $\mathcal{D}_{SEL}'$
is returned (Line 11).

\begin{algorithm}[H]
	\caption{Relative Neighborhood Graph (RNG)}
	\begin{algorithmic}[1]
		\Require {$\mathcal{D}_{SEL}$: validation set}
		\State $\mathcal{D}_{SEL}' \gets \mathcal{D}_{SEL}$
		\State $PG \gets \textit{proximity-graph}(\mathcal{D}_{SEL})$
		\For {$x_i \in \mathcal{D}_{SEL}$}
		\State $RN \gets \textit{relative-neighbors}(x_i, PG)$
		\State   \textit{label}$_{pred}$ $\gets$ most frequent class in $RN$
		\State \textit{label}$_{true}$ $\gets$ \textit{class}$(x_i)$
		\If {\textit{label}$_{true}$ $\ne$ \textit{label}$_{pred}$ $\land$ \textit{label}$_{true} \ne$ \textit{minority}$_{class}$}
		\State $\mathcal{D}_{SEL}' \gets \mathcal{D}_{SEL}'$ \textbackslash \text{ } $x_i$
		\EndIf
		\EndFor
		
		\State  \Return $\mathcal{D}_{SEL}'$
	\end{algorithmic}
	\label{alg:rng}
\end{algorithm}

\subsubsection{Edited Nearest Neighbors (ENN)}

The ENN is an edition prototype selection technique well-known
for its efficiency in removing noise and producing smoother
classes boundaries.
The ENN is used with the changes proposed
in \cite{ncr:2001}, (implemented in \cite{imblearn:2017}),
where only majority class samples are removed in order to
reduce the class imbalance.

Algorithm \ref{alg:enn} presents the pseudo-code of the ENN
technique used in this work.
Given the validation set $\mathcal{D}_{SEL}$,
all samples are added in the filtered validation set
$\mathcal{D}_{SEL}'$ (Line 1),
and for each sample $x_i \in \mathcal{D}_{SEL}$,
if $x_i$ is misclassified by its $K$ nearest neighbors
in $\mathcal{D}_{SEL}' \textbackslash x_i$ and $x_i$ is
not from the minority class,
$x_i$ is removed from the filtered validation set
$\mathcal{D}_{SEL}'$ (Line 2 - 8).
Finally, the filtered validation set $\mathcal{D}_{SEL}'$
is returned (Line 9).

\begin{algorithm}[H]
	\caption{Edited Nearest Neighbors (ENN)}
	\begin{algorithmic}[1]
		\Require {$\mathcal{D}_{SEL}$: validation set}
		\State $\mathcal{D}_{SEL}' \gets \mathcal{D}_{SEL}$
		\For {$x_i \in \mathcal{D}_{SEL}$}
		\State \textit{label}$_{pred}$ $\gets$ most frequent class in $KNN(x_i, \mathcal{D}_{SEL} \textbackslash \text{ } x_i)$
		\State \textit{label}$_{true}$ $\gets$ \textit{class}$(x_i)$
		\If {\textit{label}$_{true}$ $\ne$ \textit{label}$_{pred}$ $\land$ \textit{label}$_{true} \ne$ \textit{minority}$_{class}$}
		\State $\mathcal{D}_{SEL}' \gets \mathcal{D}_{SEL}'$ \textbackslash \text{ } $x_i$
		\EndIf
		\EndFor
		\State \Return $\mathcal{D}_{SEL}'$
	\end{algorithmic}
	\label{alg:enn}
\end{algorithm}

\subsection{Region of competence definition phase}
\label{sec:regioncompetence}

In order to solve the indecision region drawback (Section \ref{subsec:p2}), the FIRE-DES++ employs the K-Nearest Neighbors Equality (KNNE) instead of the traditional KNN algorithm in order to define the region of competence, $\Psi$, for each new query, $x_{query}$. The KNNE is a variation of the KNN technique which selects the same amount of samples from each class~\cite{knne:2011}. 

The advantage of using the KNNE instead of the original KNN method employed by the previous FIRE-DES algorithm is
that we ensure all classes are represented in the region of competence. Thus, test instances that are located close to the decision borders (i.e., in a true indecision region) will never be mistaken as belonging to a safe region. Moreover, the uses of KNNE complements the filtering stage of the FIRE-DES++ framework. By reducing the overlap between the classes, the filtering phase may remove important samples that are close to the class borders~\cite{ps-taxonomy:2012, cruz2016prototype}, which could make indecision regions being mistaken as safe regions. By using the KNNE, the FIRE-DES++ framework guarantees that the DFP mechanism will be employed in such scenarios. 

The region of competence, $\Psi$, is then passed down to the selection phase.

\subsection{Selection phase}
\label{sec:selection}

In the selection phase, first, the framework pre-selects classifiers
using the DFP. Next, a dynamic selection technique is employed, over the pre-selected pool, to select the final ensemble 
$C'$, that is used for the classification of $x_{query}$.

\subsubsection{Dynamic frienemy pruning}

\begin{figure}[H]
	\centering
	\includegraphics[scale=0.5]{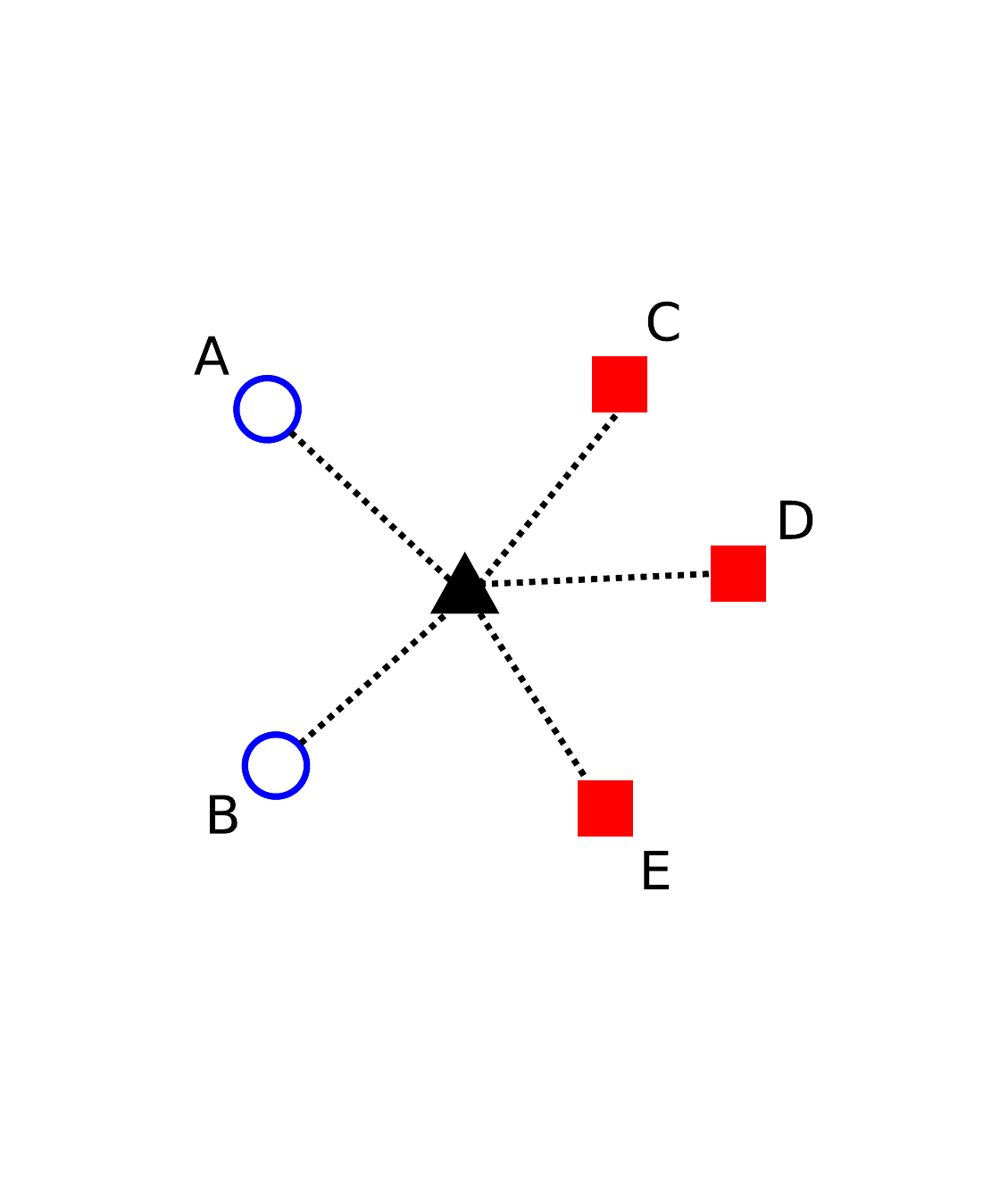}
	\caption{Pairs of frienemies $(A,C)$, $(A, D)$, $(A, E)$, $(B,C)$, $(B,D)$, $(B,E)$
		in the region of competence of the test sample $\blacktriangle$ [adapted from \cite{dfp:2017}].}
	\label{fig:frienemy}
\end{figure}

The Dynamic Frienemy Pruning (DFP) \cite{dfp:2017} aims to
pre-select competent classifiers (classifiers with decision boundaries crossing
the region of competence) for the classification of each new test sample,
before the final selection of classifiers.
The DFP algorithm uses the \textit{frienemy samples} concept:
Given a test sample $x_{query}$ and its region of competence $\Psi$,
two samples $\Psi_a$ and $\Psi_b$ are frienemy samples in regards to
$x_{query}$ if, $\Psi_a$ is in $\Psi$, $\Psi_b$ is in $\Psi$,
and $\Psi_a$ and $\Psi_b$ are from different classes.
Figure \ref{fig:frienemy} shows a test sample $\blacktriangle$
and its region of competence (samples $A$, $B$, $C$, $D$ and $E$).
In this example, the frienemy samples are the pairs of samples
of opposite classes $(\bluecircle,\redsquare)$, 
named $(A,C)$, $(A, D)$, $(A, E)$, $(B,C)$, $(B,D)$, $(B,E)$.

For each new test sample, if the test sample is located in an
indecision region, the DFP algorithm pre-selects classifiers
with decision boundaries crossing the region of competence.
That is, if the test sample have samples of different classes
in the region of competence, DFP pre-selects classifiers
that correctly classify at least one pair of frienemy samples
(if such classifier exists).

Algorithm \ref{alg:dfp} presents the DFP pseudo-code.
Given the region of competence ($\Psi$) of the test sample,
and the pool of classifiers ($C$),
DFP creates an empty list $C_{pruned}$ in which the
pre-selected classifiers will be stored (Line 1),
finds the pairs of frienemy samples ($\mathcal{F}$) in $\Psi$
(Line 2), and, for each classifier $c_i$ in $C$, $c_i$ is included
in $C_{pruned}$ if $c_i$ correctly classify at least one pair
of frienemies (Lines 3 - 8). If no classifier is pre-selected,
DFP includes all classifiers in $C$ into $C_{pruned}$ (lines 9 - 11).
Finally, $C_{pruned}$ is returned (Line 12).

\begin{algorithm}[htbp]
	\caption{Dynamic Frienemy Pruning}
	\small
	\begin{algorithmic}[1]
		\Require $\Psi$: region of competence of the test sample
		\Require $\mathcal{C}$: pool of classifiers
		\State $C_{pruned}$ $\gets $ empty ensemble of classifiers
		\State $\mathcal{F}$ $\gets $ all pair of frienemies in $\Psi$
		\For {$c_i$ in $\mathcal{C}$}
		\State $\mathcal{F}_i \gets$ pairs of samples in $\mathcal{F}$ correctly classified by $c_i$.
		\If {$\mathcal{F}_i$ is not empty}
		\State $C_{pruned} \gets C_{pruned} \cup c_i$
		\EndIf
		\EndFor
		\If {$C_{pruned}$ is empty}
		\State $C_{pruned} \gets \mathcal{C}$
		\EndIf
		\State \Return $C_{pruned}$
	\end{algorithmic}
	\label{alg:dfp}
\end{algorithm}

\subsection{Dynamic Selection}

In this step, the pruned pool $C_{pruned}$ and the region of competence, $\Psi$, are passed down to a DES technique which selects an ensemble $C'$, from $C_{pruned}$, containing the most competence base classifiers for the classification of $x_{query}$.

\begin{figure}[h]
	\begin{center}
		\includegraphics[scale=0.3]{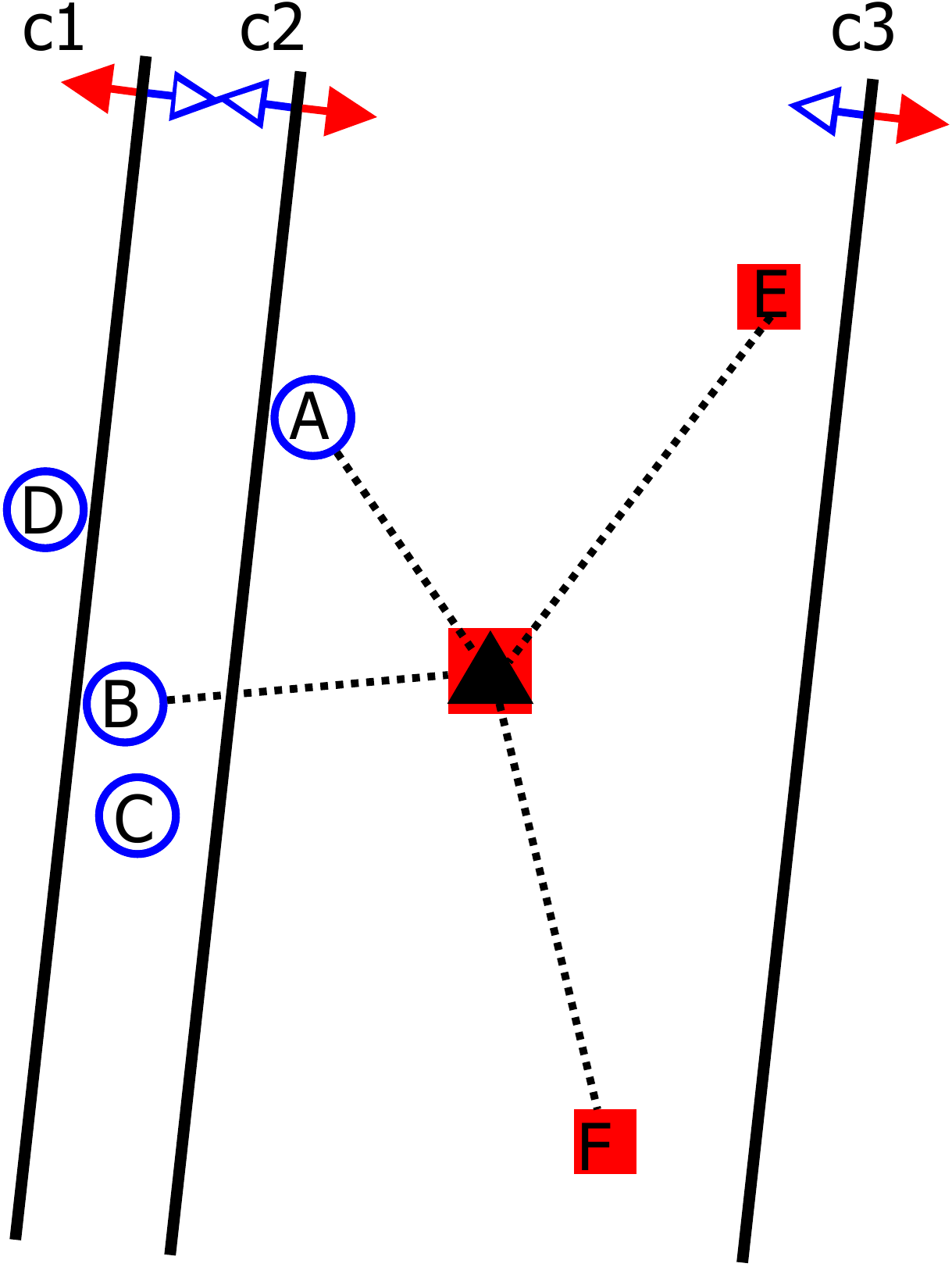}
		\caption{
			DES applied to the classification of a test sample $\blacktriangle$
			of class $\textcolor{red}{\blacksquare}$.
			The continuous straight lines are the decision boundaries of
			classifiers $c1$, $c2$, and $c3$,
			the markers {\Large$\textcolor{blue}{\circ}$} (A, B, C, and D) and
			$\textcolor{red}{\blacksquare}$ (E, and F)
			are samples of different classes,
			and samples connected to the test sample by a dotted line
			(A, B, E, and F)
			define the region of competence of the test sample.
		}
		\label{fig:ps4}
	\end{center}
\end{figure}

Figure \ref{fig:ps4} shows the same scenario from Figure \ref{fig:ps},
but without the noisy sample $N$, and using the KNNE to define the
region of competence of the test sample. First, the FIRE-DES++ removes noise from the validation set
(the example from Figure \ref{fig:ps1} is turned into the example
from Figure \ref{fig:ps2}), tackling the noise sensitivity drawback of FIRE-DES.
Then, the framework uses the KNNE to define the region of competence, 
selecting an equal amount of samples from different classes (the example
from Figure \ref{fig:ps2} is turned into the example from Figure \ref{fig:ps4}),
tackling the indecision region restriction drawback of FIRE-DES.
The region of competence now is composed of the samples A, B, E, and F (instead of A, B, C, F) 
due to the use of KNNE.

In this example,
the classifier $c1$ now correctly classifies 2 samples in $\Psi$,
the classifier $c2$ now correctly classifies 3 samples in $\Psi$,
and
the classifier $c3$ now correctly classifies 2 samples in $\Psi$. The
OLA technique now selects $c2$, correctly classifying the test sample.

By applying the DFP in this example (after the PS technique and the KNNE),
FIRE-DES++ pre-selects the classifier $c2$ as it
is the only classifier that correctly classifies at least one
pair of frienemies, correctly classifying the test sample as
being from the class $\redsquare$. In this example, FIRE-DES++
performed optimal classification for
OLA
and the same concept can be extended to other DES techniques.

\section{Experiments}
\label{sec:experiments}

In this section, we evaluate FIRE-DES++ using different dynamic selection techniques. We evaluate the impact of
the filtering phase using the PS techniques,
the region of competence definition phase using the K-Nearest Neighbors Equality (KNNE),
and the selection phase, using the Dynamic Frienemy Pruning (DFP).
We also compare the filtering phase using the ENN and RNG.

\subsection{Dynamic Selection Techniques}

We used 8 dynamic classifier selection techniques
from the literature. (Table \ref{tab:dsexp}):
Overall Local Accuracy (OLA), Local Class Accuracy (LCA), A Priori (APRI), A Posteriori (APOS),
Multiple Classifier Behavior (MCB), Dynamic Selection KNN (DSKNN), K-Nearest Oracles Union (KNU), 
and K-Nearest Oracles Eliminate (KNE). These eight techniques were selected since they are the most well-known dynamic selection techniques, having the highest number of citations according to Google Scholar. Moreover, they are all based on the KNN to estimate the region of competence. So they are suitable to be used in the FIRE-DES++ framework. A step-by-step explanation of such techniques can be found in the following surveys \cite{dcs:2014, cruz2018dynamic}.

In addition, we compare the proposed FIRE-DES++ with the three dynamic ensemble selection frameworks that achieved the best classification performance in \cite{cruz2018dynamic}: Randomized Reference Classifier (RRC) \cite{rrc:2011}, META-DES \cite{metades:2015}, and META-DES.Oracle \cite{metadesoracle:2017}. They are briefly described below:

\begin{itemize}	
	\item \textbf{RRC}: Instead of estimating the competence of the base classifiers in the neighborhood of the query, this method uses all samples in $\mathcal{D}_{sel}$, and weights the influence of each example using a Gaussian potential function so that samples closer to the query have a higher influence in the competence estimation than the more distant ones. The source of competence is estimated based on the concept of randomized reference classifier (RRC) proposed in~\cite{rrc:2011}. The base classifiers that presented a competence level higher than the random classifier are selected to compose the ensemble for an input $x_{query}$. 
	
	\item \textbf{META-DES}: The META-DES is a dynamic ensemble selection framework that model the competence estimation as a meta-problem. Each measure used to estimate the local competence of a base classifier is encoded as a meta-feature. Five sets of meta-features for the estimation of the classifier competence are considered. Then, a meta-classifier is trained, based on the training data, to predict whether or not a base classifier is competent enough for the classification of a new input $\mathbf{x}_{query}$. 
	
	\item \textbf{META-DES.Oracle}: The META-DES.Oracle is an extension of the META-DES framework based on the concept of Oracle, that is an ideal dynamic selection scheme which always selects the classifiers that predict the correct label for the current sample if such classifier exists~\cite{kuncheva:2004}. In this case, the Oracle definition is used in an optimization scheme, so that the meta-classifier can achieve results that are closer to the Oracle, improving the dynamic selection of base classifiers.
\end{itemize}

These state-of-the-art frameworks are not based exclusively on the KNN for the competence level estimation. Hence, neither the KNNE nor the DFP can be applied to these techniques.

\begin{table}[H]
	\begin{center}
		\caption{Dynamic selection techniques considered in the experiments.}
		\label{tab:dsexp}
		\begin{tabular}{lll}
			\hline
			Technique & Category & Reference \\
			\hline
			\hline
			\textbf{DCS} & &  \\
			Overall Local Accuracy (OLA)        & Accuracy      & Woods et al. \cite{dcs_la:1996} \\
			Local Class Accuracy (LCA)          & Accuracy      & Woods et al. \cite{dcs_la:1996} \\
			A Priori (APri)                     & Probabilistic &  Giacinto et al. \cite{dcs_la:1999} \\
			A Posteriori (APos)                 & Probabilistic &  Giacinto et al. \cite{dcs_la:1999} \\
			Multiple Classifier Behavior (MCB)  & Behavior      & Giacinto et al. \cite{mcb:2001} \\
			\textbf{DES} & & \\
			Dynamic Selection KNN (DSKNN)       & Diversity     & Santana et al. \cite{dsknn:2006} \\
			K-Nearests Oracles Union (KNU)      & Oracle        & Ko et al. \cite{des:2008} \\
			K-Nearests Oracles Eliminate (KNE)  & Oracle        & Ko et al. \cite{des:2008} \\
			\textbf{State-of-the-art} & & \\
			Randomized Reference Classifier (RRC)   & Probabilistic &   Woloszynski et al. \cite{rrc:2011}      \\
			META-DES                                & Meta-learning &   Cruz et al. \cite{metades:2015}         \\
			META-DES.Oracle                         & Meta-learning &   Cruz et al. \cite{metadesoracle:2017}   \\
			\hline
		\end{tabular}
	\end{center}
\end{table}

The experiments were conducted using the Python 3.5 language with the scikit-learn library~\cite{pedregosa2011scikit} for the training of the base classifiers. The dynamic ensemble selection techniques were evaluated using the DESlib library~\cite{cruz2018deslib}, which contains fast implementation of all dynamic ensemble selection techniques evaluated in this work. The library is publicly available on GitHub: \url{https://github.com/Menelau/DESlib}.

The size of the region of competence (neighborhood size) $K$ was equally set to 7 for all dynamic selection technique (as suggested in \cite{cruz2018dynamic}). This is the only hyper-parameter required for the majority of dynamic selection methods. The only exception is the DS-KNN technique, which requires to predefine the number of selected base classifiers. In this case, the number of base classifiers selected using accuracy ($N$) and diversity ($J$) was set to $30\%$ of the whole pool as suggested in~\cite{dsknn:2006}.

For the state-of-the-art techniques, the RRC has no hyper-parameter to set. The META-DES framework has two additional hyper-parameters: The number of samples selected using output profiles $K_{p}$ and the sample selection threshold $h_{c}$. The values of the hyper-parameters $K_{p}$ and $h_{c}$ for the META-DES framework were set to 5 and 80\% according to the results presented in~\cite{metades:2015, metadesoracle:2017}.

\subsection{Datasets}

\begin{table}[!t]
	\scriptsize
	\begin{center}
		\caption{
			Characteristics of the 64 datasets used in the experiments:
			label, name, number of features, number of samples, and imbalance ratio.
			The imbalance ratio (IR) is is calculated by the number of instances of the majority class per instance of the minority class.
		}
		\label{tab:datasets}
		\resizebox{!}{.50\textwidth}{
			\begin{tabular}{lrrr|lrrr}
				\hline
				
				Name            & \#Feats. & \#Samples & IR  & Name            & \#Feats. & \#Samples & IR \\
				\hline

				glass1         					 & 9     & 214        & 1.82 & ecoli-0-2-6-7vs3-5           	 & 7     & 224		  & 9.18 \\
				ecoli0vs1      					 & 7     & 220        & 1.86 & glass-0-4vs5          			 & 9     & 92		  & 9.22 \\
				wisconsin      					 & 9     & 683        & 1.86 & ecoli-0-3-4-6vs5         		 & 7     & 205		  & 9.25 \\
				pima           					 & 8     & 768        & 1.87 & ecoli-0-3-4-7vs5-6        		 & 7     & 257		  & 9.28 \\
				iris0          					 & 4     & 150        & 2.00 & 	yeast-05679vs4 					 & 8     & 528        & 9.35 \\
				glass0         					 & 9     & 214        & 2.06 &  vowel0         					 & 13    & 988        & 9.98 \\
				yeast1         					 & 8     & 1484       & 2.46 & ecoli-0-6-7vs5       			 & 6     & 220		  & 10.00 \\
				haberman						 & 3	 & 306		  & 2.78 & glass-016vs2    				 & 9     & 192        & 10.29 \\
				vehicle2       					 & 18    & 846        & 2.88 & ecoli-0-1-4-7vs2-3-5-6     		 & 7     & 336		  & 10.59 \\
				vehicle1       					 & 18    & 846        & 2.90 & led7digit-0-2-4-5-6-7-8-9vs1     & 7     & 443		  & 10.97 \\
				vehicle3       					 & 18    & 846        & 2.99 & glass-0-6vs5   					 & 9     & 205		  & 11.00 \\
				glass0123vs456 					 & 9     & 214        & 3.20 & ecoli-0-1vs5    				 & 6     & 240		  & 11.00 \\
				vehicle0       					 & 18    & 846        & 3.25 & glass-0-1-4-6vs2  				 & 9	 & 205		  & 11.06 \\
				ecoli1         					 & 7     & 336        & 3.36 & glass2          				 & 9     & 214        & 11.59 \\
				new-thyroid1   					 & 5     & 215        & 5.14 & ecoli-0-1-4-7vs5-6  			 & 6	 & 332		  & 12.28 \\
				new-thyroid2   					 & 5     & 215        & 5.14 & ecoli-0-1-4-6vs5  				 & 6	 & 280 		  & 13.00 \\
				ecoli2         					 & 7     & 336        & 5.46 & cleveland-0vs4 	  				 & 13    & 177		  & 12.62 \\
				segment0       					 & 19    & 2308       & 6.00 & shuttle-c0vsc4  				 & 9     & 1829       & 13.87 \\
				glass6         					 & 9     & 214        & 6.38 & yeast-1vs7      				 & 7     & 459        & 14.30 \\
				yeast3         					 & 8     & 1484       & 8.10 & glass4          				 & 9     & 214        & 15.47 \\
				ecoli3         					 & 7     & 336        & 8.60 & ecoli4          				 & 7     & 336        & 15.80 \\
				page-blocks0 					 & 10 	 & 5472		  & 8.79 &  page-blocks-13vs4   			 & 10 	 & 472        & 15.86\\ ecoli-0-3-4vs5 					 & 7     & 200	 	  & 9.00 	& glass-0-1-6\_vs\_5  			 & 9	 & 184        & 19.44 \\			
				yeast-2vs4     					 & 8     & 514        & 9.08 	& shuttle-c2-vs-c4    			 & 9	 & 129        & 20.50 \\		
				ecoli-0-6-7vs3-5 				 & 7     & 202		  & 9.09 	& yeast-1458vs7   				 & 8     & 693        & 22.10 \\			
				ecoli-0-2-3-4vs5  				 & 7     & 222		  & 9.10 	& glass5          				 & 9     & 214        & 22.78 \\			
				yeast-0-3-5-9vs7-8  			 & 8     & 506		  & 9.12 	& yeast-2vs8      				 & 8     & 482        & 23.10 \\			
				glass-0-1-5vs2                	 & 9     & 172		  & 9.12 	& yeast4          				 & 8     & 1484       & 28.10 \\					
				yeast-0-2-5-7-9vs3-6-8 			 & 8     & 1004		  & 9.14 	& yeast-1289vs7   				 & 8     & 947        & 30.57 \\			
				yeast-0-2-5-6vs3-7-8-9           & 8     & 1004		  & 9.14 	& yeast5          				 & 8     & 1484       & 32.73 \\		
				ecoli-0-4-6vs5             	 	 & 6     & 203		  & 9.15	& ecoli-0137vs26  				 & 7     & 281        & 39.14 \\				
				ecoli-0-1vs2-3-5            	 & 7     & 224		  & 9.17 	& yeast6          				 & 8     & 1484       & 41.40 \\			
				
				\hline
			\end{tabular}}
		\end{center}
	\end{table}
	
	We conducted the experiments on 64 datasets from the
	Knowledge Extraction based on Evolutionary Learning (KEEL)
	repository \cite{keel}. This experimental study is focused on small datasets with different
	levels of class imbalance. So, the framework is evaluated
	under a diverse set of classification problems.
	Table \ref{tab:datasets} shows the characteristics of the datasets 
	used in this experiment: label, name, number of features, 
	number of samples and the Imbalanced Ratio (IR). The IR is a common metric used by several authors~\cite{branco2016survey, diez2015diversity} to characterize the imbalanced level of a distribution. It is calculated by the number of instances of the majority class per instance of the minority class.
	
	\subsection{Evaluation}
	
	For each dataset, the experiments were carried out using a stratified 5-fold cross validation (1 fold for test and 4 folds for training). For the sake of simplicity, we use the 5-fold partitions provided in the KEEL website. Thus, making it easier to replicate the results of this paper. The process of creating the dynamic selection dataset (DSEL) was guided by the experiments conducted in~\cite{roy2018study}. Due to the low sample size, the whole training set is used for the generation of DSEL. There is an overlap between the training bootstraps and DSEL. However, due to the randomized nature of the Bagging technique as well as the application of the PS techniques its distribution is not exactly the same. Moreover, as reported by~\cite{dietrich2003decision} a small overlap between both datasets can be suitable for dealing with small sized datasets.
	
	Similar to our previous works~\cite{dfp:2017}, the pool of classifiers $C$ was composed of 100 Perceptrons generated using the Bagging technique~\cite{bagging:1996}. The training process was conducted using the scikit-learn library~\cite{pedregosa2011scikit}. The learning rate and number of iterations used for the training were set to $\alpha=0.001$ and $n_{iter} = 100$. The activation function is the Heaviside function, which predicts 0 if the sample is on one side of the hyperplane and 1 otherwise. Moreover, each Perceptron was calibrated to estimate posterior probabilities using Platt's sigmoid model~\cite{platt1999probabilistic} provided in the scikit-learn library through the CalibratedClassifierCV class.
	
	For evaluation metric, we used the Area Under the ROC Curve (AUC)
	\cite{auc:1997}. We used the AUC because this metric has been widely
	used to evaluate the performance of classifiers on imbalanced data
	\cite{cip:2013}.
	
	Furthermore, we used the Wilcoxon Signed Rank Test \cite{wilcoxon:1945} and 
	the Sign Test~\cite{sheskin2003handbook} to conduct a pairwise comparison between techniques over all datasets. These methods were used since they were suggested by~\cite{signedtest:2006, benavoli2016should}. The Wilcoxon Signed Rank Test is a non-parametric alternative to the paired t-test. The Sign test works upon the number of wins, ties and losses obtained by an algorithm over the baseline. The algorithm is deemed statistically better if its number of wins plus half of the number of ties is higher than a critical value.
	
	Comparison between multiple techniques over all datasets is conducted using the Friedman test with the Bonferroni-dunn post-hoc test as suggested by Demsar \cite{signedtest:2006}. The Friedman test is a non-parametric equivalent of the repeated-measures ANOVA. It ranks the algorithms for each data set separately, the best one getting the rank of 1, the second best rank 2 and so on. In case of a tie, i.e., two methods presented the same classification accuracy for the dataset, their average ranks were summed and divided by two. However, the Friedman test only tells that there is a difference between the classifiers, but does not present which methods differ. For this reason, the  Bonferroni-dunn post-hoc test is employed to find out which techniques actually differs.
	
	\subsection{Filtering Phase: RNG vs. ENN}
	
	In this section, we evaluate FIRE-DES++ using RNG and ENN for the filtering phase. Both techniques follow
	the same approach of maintaining all samples of the minority class. In other words, a sample is only considered a noise and removed if it belongs to the majority class. This comparison is important for verifying whether the FIRE-DES++ is sensitive to changes in PS techniques in the filtering phase, and also for finding the PS technique
	that causes the highest classification performance gain in FIRE-DES++.
	
	\begin{figure}[H]
		\centering
		\includegraphics[scale=0.4]{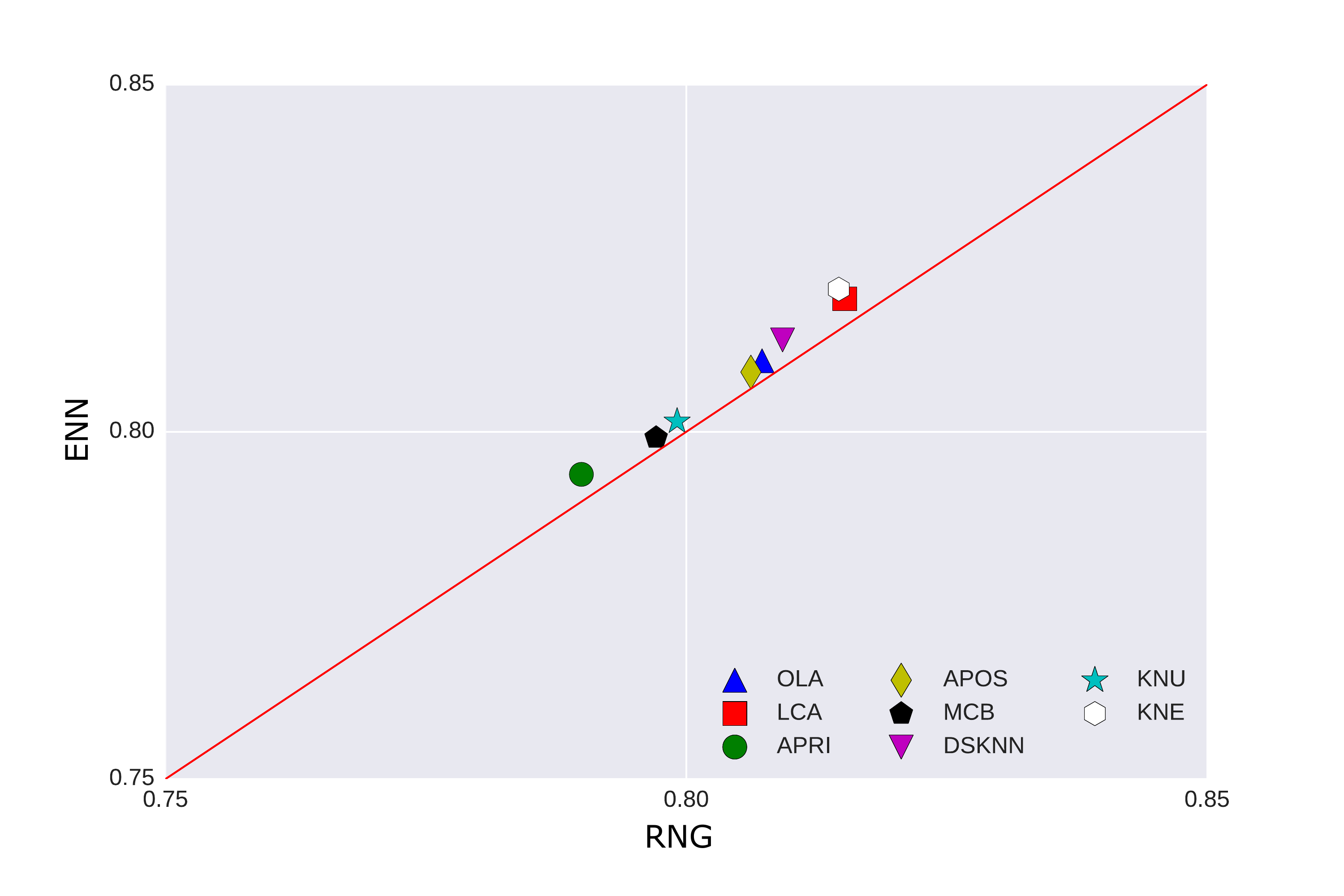}
		\caption{
			Scatter plots of average AUC of FIRE-DES++ using the ENN (vertical axis)
			and the RNG (horizontal axis). Markers above the diagonal line indicates that
			the using the ENN had a better performance than using the RNG.
		}
		\label{fig:scatter_ff}
	\end{figure}
	
	Figure \ref{fig:scatter_ff} shows the scatter plot of average AUC of FIRE-DES++
	using the ENN (vertical axis) and the RNG (horizontal axis). In this figure, all markers
	are above the diagonal line, meaning that using the ENN was, on average, better than
	using the RNG for all DES techniques in the proposed framework.
	
	Using the Wilcoxon Signed Rank Test ($\alpha = 0.05$), we can confirm that
	using the proposed framework with the ENN is statistically better than RNG for the majority of DES techniques:
	OLA (\textit{p-value} = $0.0121$),
	LCA (\textit{p-value} = $0.0011$),
	APRI (\textit{p-value} = $0.0040$),
	MCB (\textit{p-value} = $0.0007$),
	DSKNN (\textit{p-value} = $0.0002$),
	KNU (\textit{p-value} = $0.0010$),
	and
	KNE (\textit{p-value} = $0.0002$). The only exception is for the APOS technique (\textit{p-value} = $0.0946$).
	Thus, we only consider FIRE-DES++ using ENN for the rest of this paper.
	
	\subsection{Comparison among different scenarios}
	\label{comp:scenarios}
	
	In this section, we analyze eight different scenarios for the dynamic selection techniques (Table 3). Each Scenario corresponds to a different combination of the three modules present in the FIRE-DES++ framework: DFP, ENN, and KNNE. Scenario I corresponds to the original dynamic selection techniques (i.e., no additional step is performed). Scenario IV corresponds to the FIRE-DES framework, in which only the DFP method is applied without using the modifications proposed in this paper (ENN and KNNE). Scenario VIII corresponds to the FIRE-DES++, in which the DFP, ENN and KNNE are all employed in the framework.
	
	\begin{table}[H]
		\scriptsize
		\begin{center}
			\caption{
				Eight test scenarios considered this work. Scenarios I, IV and VIII corresponds to the standard DES techniques, the FIRE-DES framework and FIRE-DES++ framework respectively.
			}
			\label{tab:scenarios}
			\begin{tabular}{lrrr}
				\hline
				Scenario    &   KNNE &   ENN    &   DFP \\
				\hline
				\hline  
				I       &   No  &   No  &   No  \\
				II      &   Yes &   No  &   No  \\
				III     &   No  &   Yes &   No  \\
				IV      &   No  &   No  &   Yes \\
				V       &   Yes &   Yes &   No  \\
				VI      &   Yes &   No  &   Yes \\
				VII     &   No  &   Yes &   Yes \\
				VIII    &   Yes &   Yes &   Yes \\
				\hline
			\end{tabular}
		\end{center}
	\end{table}
	
\begin{figure}[H]
		\centering
		\includegraphics[scale=0.7]{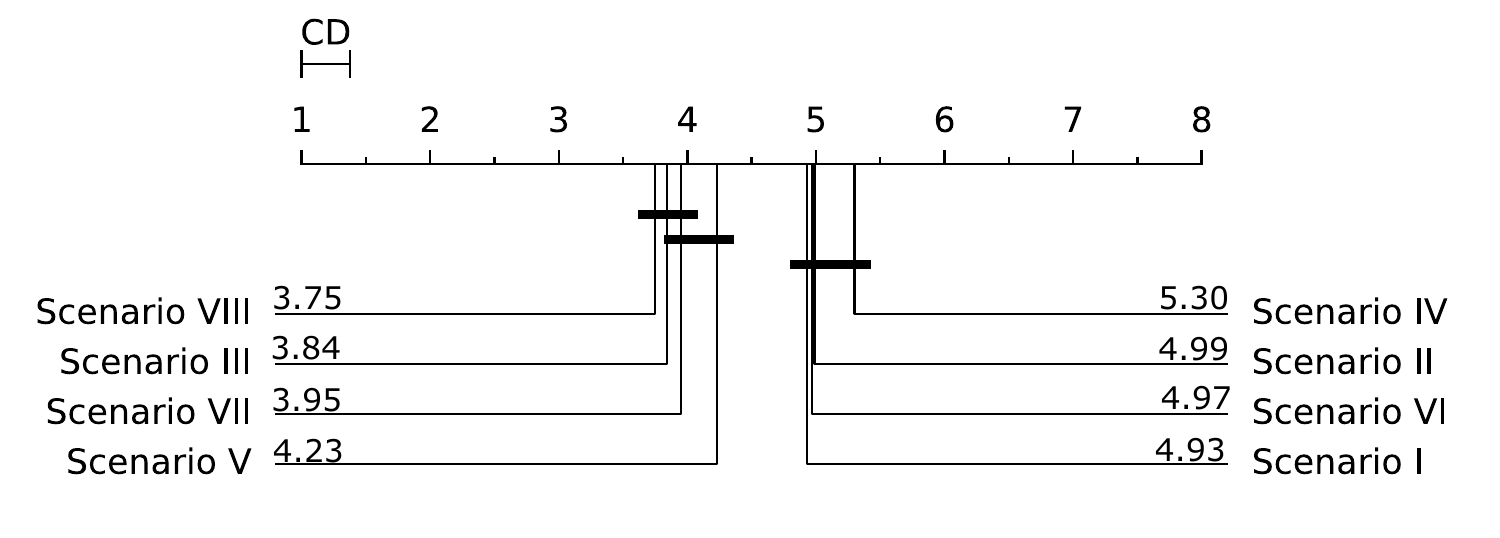}
		\caption{
			Critical Difference diagram using the Bonferroni-dunn post-hoc test considering the eight Scenarios. Scenarios that are statistically equivalent are connected by a black bar.
		}
		\label{fig:diagram_scenarios}
	\end{figure}

	For each scenario, we evaluated the classification performance
	of each DES technique over the 64 datasets, a total of 512 experiments
	(64 datasets $\times$ 8 DS techniques) per scenario. We performed the Friedman test to have a comparison between the eight scenarios considering all datasets. For
	each dataset and dynamic selection technique, we ranked each scenario from rank 1 to rank 8 (rank 1 being the best),
	and used the Friedman test to calculate their average rank  (Table~\ref{table:aucScenarios}).
	The result of the Friedman test was
	$\text{\textit{p-value}} = 2.39 \times e^{-70}$, indicating that there is statistical difference
	between the scenarios. In order to know where the difference lies, the Bonferroni-Dunn post-hoc test is conducted. The result of the post-hoc analysis is presented using a critical difference diagram (Figure \ref{fig:diagram_scenarios}). Scenarios significantly different have a difference in ranking higher than the critical difference ($CD = 0.3750$).
	
	
	\begin{table*}[h!] 
		\centering 
		\caption{The average ranks and AUC for each Scenario. The Scenarios are ordered according to their performance.} 
		\label{table:aucScenarios}  
		\resizebox{0.60\textwidth}{!}{  
			\begin{tabular}{|l c || l c|}  
				\hline  
				
				Algorithm & Avg. Rank & Algorithm & Mean AUC \\ 			 
				
				\hline
				Scenario VIII & 3.75 & Scenario VIII & 82.95 \\
				Scenario III & 3.84	& Scenario VII & 82.70 \\ 								
				Scenario VII & 3.95 & Scenario III & 82.13 \\ 
				Scenario V & 4.23 & Scenario V & 82.11 \\
				Scenario I & 4.93 &  Scenario VI & 81.57 \\
				Scenario VI & 4.97 & Scenario II & 81.37 \\
				Scenario II & 4.99 & Scenario IV & 81.18 \\
				Scenario IV & 5.30 & Scenario I & 80.61 \\
				
				\hline  
				
			\end{tabular}} 
		\end{table*}

		Figure \ref{fig:diagram_scenarios} shows that FIRE-DES++ (Scenario VIII) achieved the lowest average ranking ($3.75$), statistically
		outperforming Scenarios I, II, IV, V, and VI. Scenarios VI (DFP+KNNE) and VII (DFP+ENN) obtained lower average rank when compared to scenario IV (DFP alone). 		
		The reason for Scenario IV obtaining the highest average rank in this analysis is due to the fact that it never obtained the best result (lowest rank) for any combination of 64 datasets $\times$ 8 DES techniques. There is always a better alternative either by using DFP+ENN to solve the noise sensitivity drawback (Section~\ref{subsec:p1}), DFP+KNNE to solve the indecision region definition drawback (Section 2.3~\ref{subsec:p2}) or using them all together. Thus, we can conclude the addition of ENN and KNNE really helps in improving the performance of the FIRE-DES framework.
		
		\begin{figure}[!hbtp]
			\centering
			\includegraphics[scale=0.45]{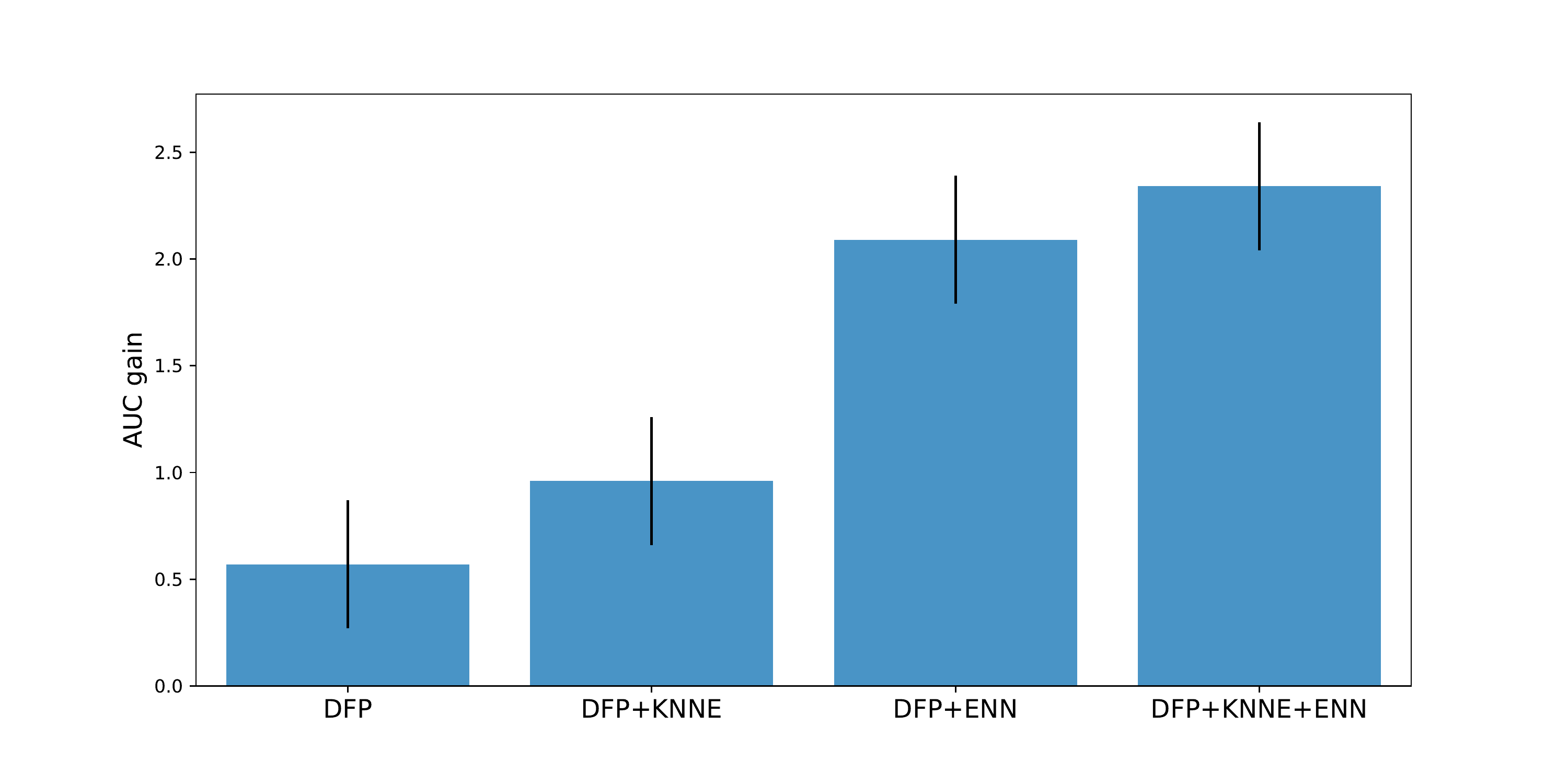}
			\caption{
				Influence of each phase when compared to Scenario I,
				that is, the difference between the average performance Scenarios IV, VI, VII and VIII
				in relation to Scenario I.
				The bars represent average classification performance gain (AUC)   
				when adding DFP (0.57), DFP+KNNE (0.96),
				DFP+ENN (2.09), and DFP+KNNE+ENN (2.34),  over the 64  datasets.
			}
			\label{fig:effects}
		\end{figure}

		Figure \ref{fig:effects} shows the performance gain (AUC) obtained by adding each step of the proposed FIRE-DES++ framework
		in relation to the regular DES techniques. The regular DES techniques corresponds to Scenario I (Table~\ref{tab:scenarios}), while the DFP, DFP+KNNE, DFP+ENN, and DFP+KNNE+ENN corresponds to Scenarios IV, VI, VII, and VIII respectively.
		This figure shows that the three phases combined
		(DFP, KNNE, and ENN) causes the highest performance
		gain (2.34), followed by DFP and ENN combined (2.09),
		DFP and KNNE combined (0.96), and finally DFP alone (0.57).
		These results indicate that the filtering and the region of competence
		definition phases in the FIRE-DES++ framework cause performance gain
		over FIRE-DES, with the performance best being the use of both the ENN and KNNE combined.
		
		Thus, we can conclude that all steps of FIRE-DES++ are important. Each step helps in improving the performance of the DES techniques. Furthermore, using all three combined leads to the highest overall improvement in
		classification performance.

		\subsection{Comparison with FIRE-DES}
		\label{comp:des}
		
		In this section, we compare FIRE-DES++ and FIRE-DES for each DES technique
		considered in this work. The goal of this analysis is to investigate whether FIRE-DES++
		significantly improves the performance of FIRE-DES as well as to identify which DES techniques are more benefited from the	proposed framework. 
		
\begin{figure}[H]
			\centering
			\includegraphics[scale=0.7]{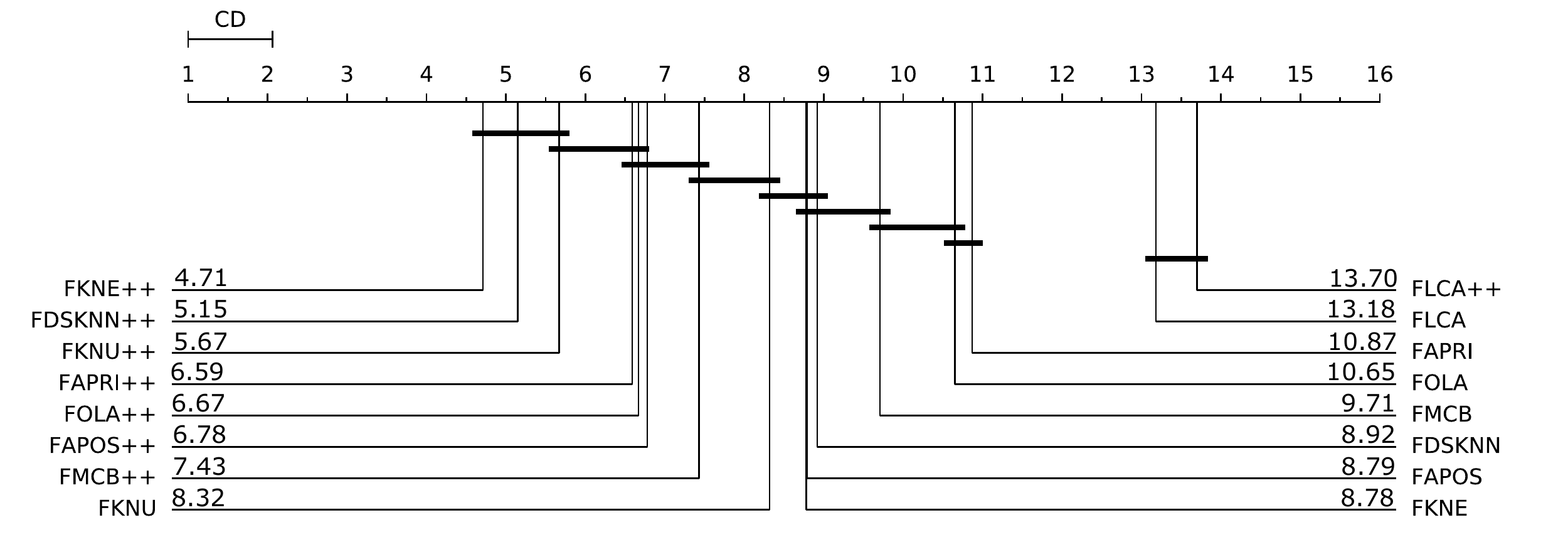}
			\caption{
				CD diagram of Bonferroni-Dunn post-hoc test considering all dynamic selection approaches. 
				$CD = 1.0608$.}
			\label{fig:diagram_all_des}
		\end{figure}

		The average rank and AUC for each DES techniques is shown on Table~\ref{table:resultsFIREDES++}. Figure \ref{fig:diagram_all_des} presents the CD diagram comparing
		FIRE-DES++ (FOLA++, FLCA++, FAPRI++, FAPOS++, FMCB++, FDSKNN++, FKNU++, and FKNE++)
		with FIRE-DES (FOLA, FLCA, FAPRI, FAPOS, FMCB, FDSKNN, and FKNE) using the Bonferroni-Dunn post-hoc test. We can see that FIRE-DES++ outperformed FIRE-DES for 7 out of 8 DES techniques. The only exception was for the LCA method, in which the FLCA and FLCA++ had statistically equivalent results.
		
\begin{figure}[H]
			\centering
			\includegraphics[scale=0.45]{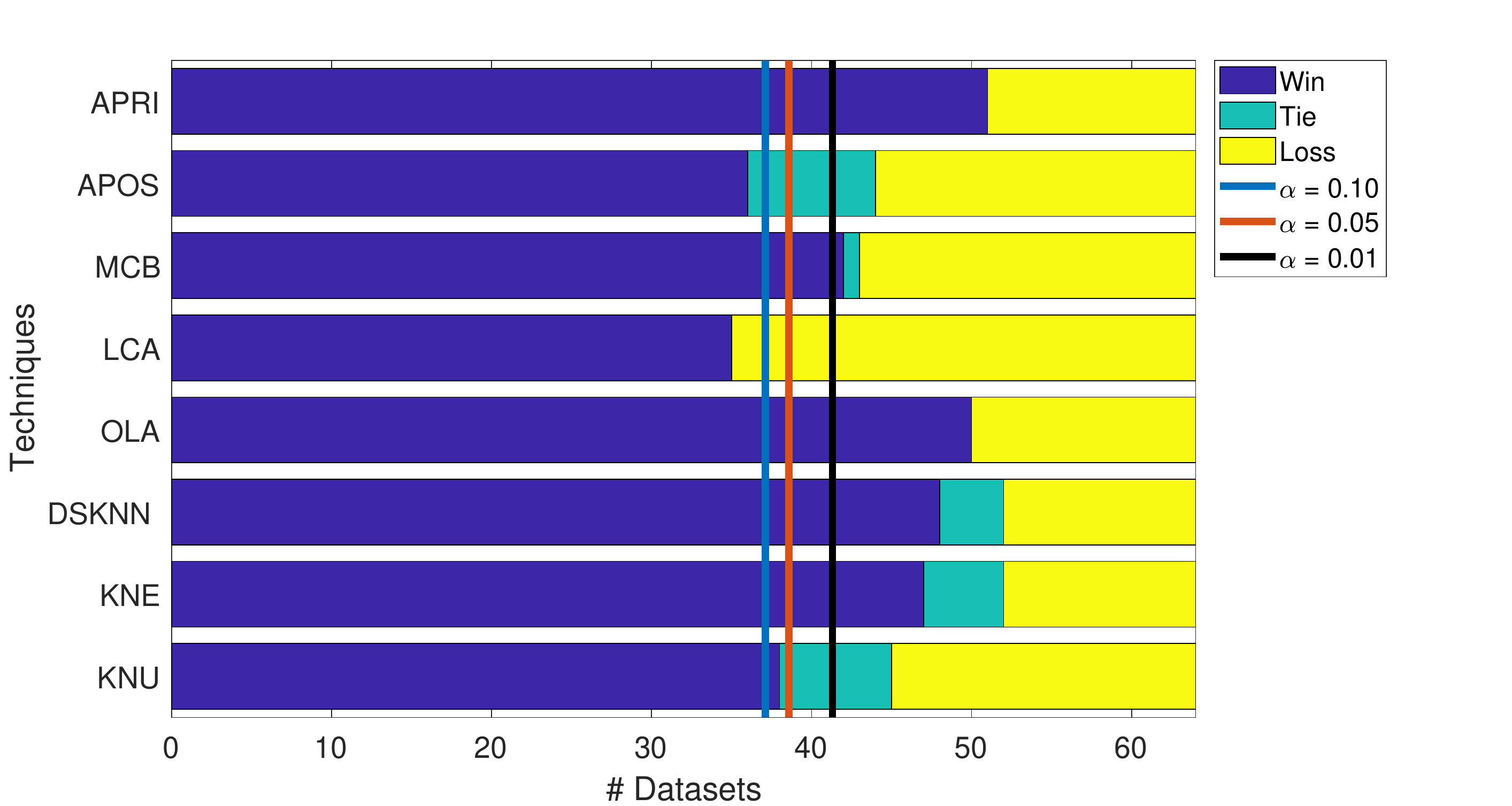}
			\caption{
				Performance of FIRE-DES++ compared with FIRE-DES using different DES techniques
				in terms of wins, ties and losses considering the average AUC over the 64 datasets.
				each line illustrates the critical values
				$n_c = \{37.12, 38.58, 41.30\}$ considering significance levels of
				$\alpha = \{0.10, 0.05, 0.01\}$, respectively.
			}
			\label{fig:wtl}
		\end{figure}

		In addition, Figure \ref{fig:wtl} presents a pairwise comparison of FIRE-DES++ and FIRE-DES for each DES technique.
		This comparison used the Sign test calculated on the computed wins, ties and losses of FIRE-DES++.
		The null hypothesis $H_0$ was that using the FIRE-DES++ did not make any difference compared to FIRE-DES,
		and a rejection of $H_0$ meant that FIRE-DES++ significantly outperformed FIRE-DES.
		In this evaluation, we considered three levels of significance $\alpha = \{0.10, 0.05, 0.01\}$.
		To reject $H_0$, the number of wins plus half of the number of ties
		needs to be greater or equal to a critical value $n_c$ (Equation \ref{eq:ncp}):
		
		\begin{equation}
		\label{eq:ncp}
		n_c = \dfrac{n_{exp}}{2} + z_{\alpha} \times \dfrac{\sqrt[2]{n_{exp}}}{2}
		\end{equation}
		
		\noindent where $n_{exp} = 64$ (the number of experiments),
		$n_c = \{37.12, 38.58, 41.30\}$ is the critical value for each significance level $\alpha = \{0.10, 0.05, 0.01\}$,  respectively.
		
		\begin{table*}[h!] 
			\centering 
			\caption{Overall results considering the 64 datasets. The average ranks and AUC for each algorithm is presented. The algorithms are ordered according to their performance} 
			\label{table:resultsFIREDES++}  
			\resizebox{0.60\textwidth}{!}{  
				\begin{tabular}{|l c || l c|}  
					\hline  
					
					Algorithm & Avg. Rank & Algorithm & Mean AUC \\ 			 
					
					\hline
					FKNE++ 		& 4.71  & FKNE++ & 85.17 \\
					FDSKNN++ 	& 5.15 	& FDSKNN++ & 85.02 \\ 								
					FKNU++		& 5.67  & FOLA++ & 84.35 \\ 
					FAPRI++		& 6.59  & FAPRI++ & 84.23 \\
					FOLA++		& 6.67  & FKNU++ & 84.22 \\
					FAPOS++		& 6.78  & FMCB++ & 83.95 \\
					FMCB++		& 7.43  & FAPOS++ & 83.66 \\
					FKNU			& 8.32  & FKNU & 82.69 \\
					
					FKNE			& 8.78  & FAPOS & 82.59 \\			
					FAPOS		& 8.79  & FKNE & 82.25 \\
					FDSKNN		& 8.92  & FMCB & 81.92 \\
					FMCB			& 9.71  & FDSKNN & 81.87  \\
					FOLA			& 10.65 & FOLA & 81.46 \\
					FAPRI		& 10.87 & FAPRI & 81.39 \\
					FLCA			& 13.18 & FLCA  & 77.66 \\			
					FLCA++		& 13.70 & FLCA++ & 77.50 \\
					
					\hline  
					
				\end{tabular}} 
			\end{table*} 
			
			Figure \ref{fig:wtl} shows that FIRE-DES++ caused a significant performance gain over FIRE-DES based on the Sign test. For a confidence level $\alpha = \{0.10, 0.05\}$ (first 2 lines left to right), FIRE-DES++ significantly improved the performance of 7 out of 8 techniques. In addition, with a more restrict confidence level $\alpha = 0.01$, the proposed FIRE-DES++ presented statistically better results for the A Priori, A Posteriori, MCB, OLA, DSKNN and KNE.
			Only for the LCA technique the FIRE-DES++ did not significantly improve over the FIRE-DES framework. However, the FLCA++ still obtained a higher number of wins (35) than losses (29). Thus, we can conclude that by the addition of ENN filter and the KNNE, the FIRE-DES++ can significantly improve the performance of a diverse set of dynamic selection techniques.
			
			In addition, we measured the processing time of the original FIRE-DES framework and the proposed FIRE-DES++ framework. The processing time was calculated by computing the average processing time over the 64 datasets. The average running time of the proposed FIRE-DES++ framework was about 10\% slower than the original FIRE-DES framework. Therefore, we can conclude that the FIRE-DES++ significantly improves the performance of DES techniques with a minimal increase in the computational time.

			\subsection{Comparison with state-of-the-art}
			\label{comp:state_of_the_art}
			
			In this section we compare the results of the FIRE-DES++ with the state-of-the-art dynamic ensemble selection frameworks (Table \ref{tab:dsexp}) as well as static ensemble methods. The following static ensemble methods were considered: Bagging \cite{bagging:1996}, AdaBoost \cite{freund1995desicion}, Random Forests \cite{breiman2001random}, Extremely Randomized Forest \cite{geurts2006extremely}, Gradient Boosted Trees \cite{friedman2002stochastic} and Random Balance ensembles \cite{diez2015random}. Each technique was evaluated with a total of 100 base classifiers. The hyper-parameters of such techniques were set with the values suggested in \cite{delgado14a}.
			
			For the sake of simplicity, only the FKNE++ was considered in this analysis since it performed better in the previous experiments. Table \ref{table:auc_state_of_the_art} presents the average AUC and ranking of FKNE++, 
			the state-of-the-art DES frameworks and the static ensemble methods. The FKNE++ obtained the lowest average rank (2.84), and the second best average AUC, 85.17 vs 85.37 obtained by the Random Balance ensemble.
			
			\begin{table*}[h!] 
				\centering 
				\caption{Overall results considering the 64 datasets. The average ranks and AUC for each algorithm is presented. The algorithms are ordered according to their performance} 
				\label{table:auc_state_of_the_art}  
				\resizebox{0.60\textwidth}{!}{  
					\begin{tabular}{|l c || l c|}  
						\hline  
						
						Algorithm & Avg. Rank & Algorithm & Mean AUC \\ 			 
						
						\hline
						FKNE++ & 2.84 & Random Balance & 85.37 \\
						Random Balance 	& 3.32 	& FKNE++ & 85.17 \\ 								
						META-DES.O & 4.84 & META-DES.O & 82.56 \\ 
						Boosting & 5.09 & META-DES & 82.18 \\
						META-DES & 5.20 & Gradient Boosting & 81.00 \\
						Gradient Boosting & 5.43 & Boosting & 80.76 \\
						RRC & 5.48 & RRC & 80.50 \\
						Bagging & 5.93 & Bagging & 78.41 \\
						Extreme Forest & 6.68 & Extreme Forest & 78.00 \\
						
						\hline  
						
					\end{tabular}} 
				\end{table*} 
				
				\begin{figure}[h]
					\centering
					\includegraphics[scale=0.55]{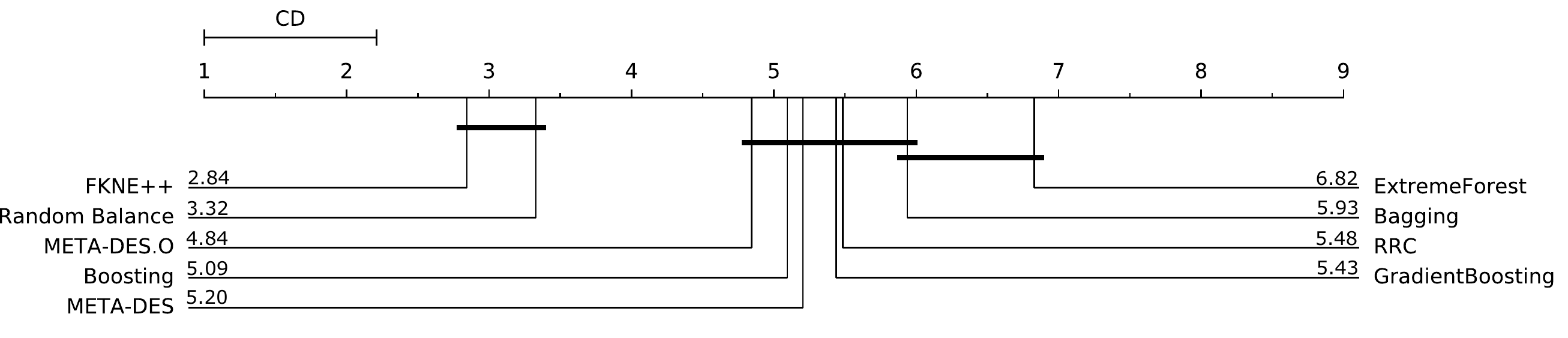}
					\caption{
						Critical difference diagram of Bonferroni-Dunn post-hoc test considering the state-of-the-art DES frameworks and  static ensemble approaches. The critical value was computed using a confidence level $\alpha=0.05$ ($CD = 1.2028$).}
					\label{fig:diag_state_of_the_art}
				\end{figure}

				Moreover, Figure~\ref{fig:diag_state_of_the_art} presents the results of the rank analysis using critical difference diagram. The critical value was computed using the Bonferroni-Dunn test with a confidence level $\alpha=0.05$ ($CD = 1.2028$).
				We can see that the FKNE++ statistically outperformed all state-of-the-art DES framework based on the rank analysis. Using the Wilcoxon Signed Rank Test ($\alpha=0.05$) for a more robust pairwise analysis, we also observed that FKNE++ statistically outperformed all three state-of-the-art DES frameworks: META-DES (\textit{p-value} $= 1.29  \times e^{-6}$), META-DES.Oracle (\textit{p-value} $= 2.95 \times e^{-5}$) and RRC (\textit{p-value} $= 2.33 \times e^{-6}$). Thus, we can conclude the proposed FIRE-DES++ presents a significant performance gain over the state-of-the-art DES frameworks for these datasets.
				
				The FKNE++ also statistically outperformed the majority of static ensemble combination methods. The only exception being the Random Balance technique. This could be explained by the fact the Random Balance was proposed to deal specifically with small sized and imbalanced data~\cite{diez2015random}, which comprises the 64 datasets in this study. Moreover, this technique achieved the state-of-the-art performance for such datasets in several comparative studies \cite{diez2015diversity, roy2018study}. Hence, the FKNE++ is competitive with the state-of-the-art methods for dealing with small sized and imbalanced datasets.

\section{Conclusion}
\label{sec:conclusion}

In this paper, we presented 2 drawbacks of the
Frienemy Indecision REgion Dynamic Ensemble Selection (FIRE-DES)
framework:
(1) noise sensitivity drawback:
the classification performance of FIRE-DES is strongly affected
by noise, as it mistakes noisy regions for indecision
regions and applies the pre-selection of classifiers.
(2) indecision region restriction drawback:
FIRE-DES uses the region of competence to decide if a test sample
is located in an indecision region, and only pre-selects classifiers
when the region of competence of the test sample is composed of
samples from different classes, restricting the number of test samples
in which the pre-selection is applied for its classification.

To tackle these drawbacks of FIRE-DES, we use the Edited Nearest Neighbors (ENN) \cite{enn:1972}
to remove noise from the validation set (tackling the noise sensitivity drawback),
and we use the K-Nearest Neighbors Equality (KNNE) \cite{knne:2011} to define the region of competence
selecting the nearest neighbors from each class (tackling the indecision region restriction drawback).
We named this new framework FIRE-DES++.

We compared the results FIRE-DES++ with DES and FIRE-DES
with 8 dynamic selection techniques over 64 datasets.
The experimental results show that the FIRE-DES++ significantly outperform FIRE-DES
for 7 out of 8 DES techniques. Moreover, results also show that each individual phase 
of the new framework, filtering and region of competence definition, helps in significantly improving
generalization performance of DES techniques.

We also compared the performance of the FIRE-DES++ with the state-of-the-art DES frameworks and ensemble methods. 
The results showed that the proposed framework significantly outperformed all three state-of-the-art DES frameworks with
statistical confidence as well as the majority of the state-of-the-art ensemble methods. Furthermore,
the FIRE-DES++ is equivalent to the Random Balance method which is considered the state-of-the-art ensemble algorithm for dealing with the KEEL imbalanced datasets according to~\cite{diez2015diversity}. 

Future works on this topic will involve extending the FIRE-DES++ framework for handling multi-class classification problems;
evaluating the use of different types of base classifier as well as other ensemble generation methods in the framework, and performing a complete study on the FIRE-DES++ together with data preprocessing techniques for dealing with imbalanced classification problems.

\section*{Acknowledgments}

The authors would like to thank
CAPES (Coordena\c{c}\~ao de Aperfei\c{c}oamento de Pessoal de N\'ivel Superior, in portuguese),
CNPq (Conselho Nacional de Desenvolvimento Cient\'ifico e Tecnol\'ogico, in portuguese)  and
FACEPE (Funda\c{c}\~ao de Amparo \`a Ci\^encia e Tecnologia do Estado de Pernambuco, in portuguese).

\bibliographystyle{elsarticle-num}
\bibliography{e2f_bibliography}

\end{document}